\documentclass{article} 
\usepackage{iclr2023_conference,times}

\usepackage{amsmath,amsfonts,bm}

\def\eqref#1{equation~\ref{#1}}

\def\1{\bm{1}}

\def\vf{{\bm{f}}}

\def\vr{{\bm{r}}}

\def\vx{{\bm{x}}}
\def\vy{{\bm{y}}}

\def\mS{{\bm{S}}}

\DeclareMathAlphabet{\mathsfit}{\encodingdefault}{\sfdefault}{m}{sl}
\SetMathAlphabet{\mathsfit}{bold}{\encodingdefault}{\sfdefault}{bx}{n}

\DeclareMathOperator*{\argmax}{arg\,max}

\DeclareMathOperator{\sign}{sign}

\usepackage[utf8]{inputenc} 
\usepackage[T1]{fontenc}    
\usepackage{hyperref}       
\usepackage{url}            
\usepackage{booktabs}       
\usepackage{amsfonts}       
\usepackage{nicefrac}       
\usepackage{microtype}      
\usepackage{xcolor}         

\usepackage[textsize=footnotesize]{todonotes}

\usepackage{graphicx}
\usepackage{amsmath}
\usepackage{amssymb}
\usepackage{comment}
\usepackage{wrapfig}
\usepackage{multirow}
\usepackage{makecell}
\usepackage{subcaption}
\usepackage{colortbl}
\usepackage{enumitem}
\usepackage{fancyvrb}
\usepackage{color}
\usepackage{framed}

\usepackage{amssymb}
\usepackage{pifont}
\newcommand{\cmark}{\ding{51}}%
\newcommand{\xmark}{\ding{55}}%

\usepackage{hyperref}
\usepackage{url}

\title{On Trace of PGD-Like Adversarial Attacks}

\author{Mo Zhou\\
Johns Hopkins University\\
\And
Vishal M. Patel\\
Johns Hopkins University\\
}

\iclrfinalcopy 
\begin{document}

\maketitle

\begin{abstract}
    Adversarial attacks pose safety and security concerns to deep learning
	applications, but their characteristics are under-explored.
    Yet largely imperceptible, a strong trace could have been left by PGD-like attacks
    in an adversarial example.
    Recall that PGD-like attacks trigger the ``local linearity'' of a network,
	which implies different extents of linearity for benign or adversarial examples.
    Inspired by this, we construct an \emph{Adversarial Response Characteristics} (ARC)
    feature to reflect the model's gradient consistency around the input to
    indicate the extent of linearity.
    Under certain conditions, it qualitatively shows a gradually varying pattern
	from benign example to adversarial example, as the latter leads to \emph{Sequel
	Attack Effect} (SAE).
	To quantitatively evaluate the effectiveness of ARC, we conduct experiments
	on CIFAR-10 and ImageNet for attack detection and attack type recognition
	in a challenging setting.
	The results suggest that SAE is an effective and unique trace of PGD-like
	attacks reflected through the ARC feature.
	The ARC feature is intuitive, light-weighted, non-intrusive, and data-undemanding.
\end{abstract}


\section{Introduction}
\label{sec:1}

Recent studies reveal the vulnerabilities of deep neural networks by 
adversarial attacks~\citep{bim,madry}, where undesired outputs (\emph{e.g.},
misclassification) are triggered
by an imperceptible perturbation superimposed on the model input.
The attacks pose safety and security concerns for respective applications.
The PGD-like attacks, including BIM~\citep{bim}, PGD~\citep{madry}, MIM~\citep{mim},
and APGD~\citep{autoattack}, are strong and widely used in the literature,
but their characteristics are under-explored.

Yet, we speculate that a strong attack leaves a strong trace in its result,
as in the feature maps~\citep{featuredenoise}.
In this paper, we consider an \emph{extremely limited setting} -- to identify the trace
of PGD-like attacks,
given an \emph{already trained}
deep neural network and merely a \emph{tiny} set (\emph{e.g.}, $50$) of
training data, \emph{without} any change in architecture or weights, \emph{nor}
any auxiliary deep networks.
A method effective in such a setting would be less dependent on external
models or data, and tends to reveal deeper characteristics of adversarial examples.
Expectedly, it is feasible in more scenarios, even when full access to data is impossible,
such as Federated Learning~\citep{fedavg} and third-party forensics.
Meanwhile, it helps us to understand adversarial examples better.

Recall that FGSM~\citep{fgsm}, the foundation of PGD-like attacks, attributes
the network vulnerability to ``local linearity'' being easily triggered by
adversarial perturbations.
Thus, we conjecture that a network behaves in a greater extent of linearity to
adversarial examples than benign (\emph{i.e.}, unperturbed) ones. 
With the first-order Taylor expansion of a network, ``local linearity'' implies
high gradient proximity in the respective local area.
Thus, we can select a series of data points with stable patterns near the input
as exploitation vectors using the BIM~\citep{bim} attack, and then compute the
model's Jacobian matrices with respect to them.
Next, the \emph{Adversarial Response Characteristics} (ARC) matrix is
constructed from these Jacobian matrices reflecting the gradient direction
consistency across all exploitation vectors.
Unlike benign examples, results of PGD-like attacks trigger \emph{Sequel
Attack Effect} (SAE), leaving higher values in the ARC matrix, reflecting
higher gradient consistency around the input.
Visualization results suggest SAE is a gradually varying pattern with
perturbation magnitude increasing, indicating its effectiveness.

\begin{figure}[t]
	\includegraphics[width=\linewidth]{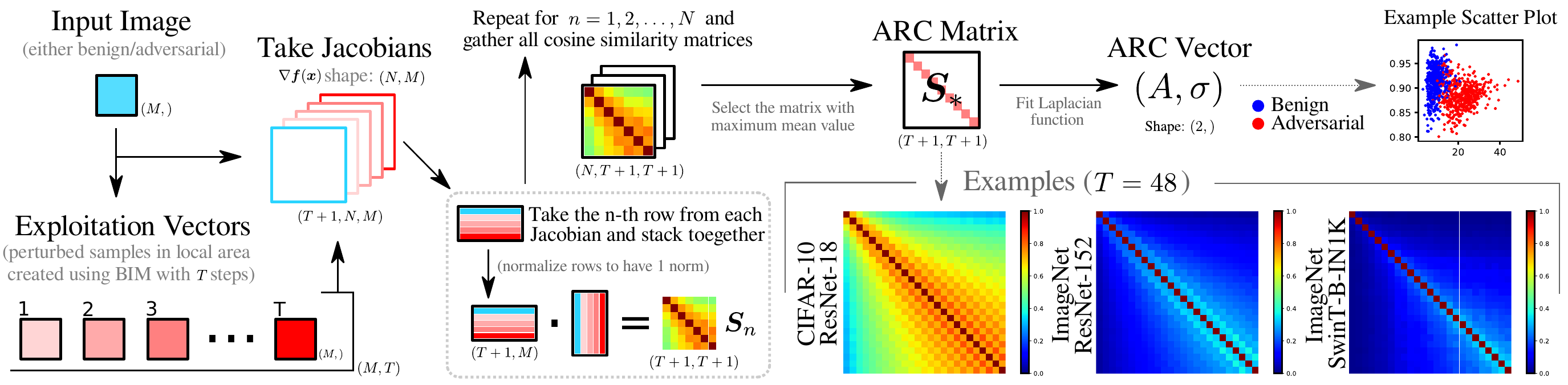}
	\caption{Diagram for computing the ARC matrix and the ARC vector.
	They reflect the model's gradient consistency within a local linear
	area around the input to indicate the extent of linearity.
	Shallow network like ResNet-18 shows higher linearity to benign examples,
	while deeper networks like ResNet-152 and SwinT-B-IN1K show lower linearity.
	%
	%
	}
    \label{fig:diag}
\end{figure}

The ARC matrix can be simplified into a $2$-D ARC vector by fitting a Laplacian
function due to their resemblance. This simplifies the interpretation of subsequent procedures.
The ARC vector can be used for \emph{informed} attack detection (the
perturbation magnitude $\varepsilon$ is known) with an SVM-based binary
classifier,
or \emph{uninformed} attack detection (the perturbation magnitude
$\varepsilon$ is unknown) with an SVM-based ordinal regression model.
The ARC vector can also be used for \emph{attack type recognition}
in similar settings with the same set of SVMs.
The SAE reflected through ARC is the unique trace of PGD-like attacks.
Due to the uniqueness of SAE to PGD-like attacks,
we can also infer attack details, including the loss function and the
ground-truth label once the attack is detected.

We evaluate our method on CIFAR-10~\citep{cifar} with ResNet-18~\citep{resnet},
and ImageNet~\citep{imagenet} with ResNet-152~\citep{resnet} / SwinT-B-IN1K~\citep{swint}.
Qualitative visualizations and quantitative experimental results for
attack detection and attack type recognition manifest the effectiveness of
our method in identifying SAE.
SAE is the unique trace of PGD-like attacks,
which also possess considerable generalization capability
among PGD-like attacks even if
training data only involves a few benign and adversarial examples.

\textbf{Contributions.}
We present the ARC features to identify the unique trace, \emph{i.e.},
SAE of PGD-like attacks from adversarially perturbed inputs.
It can be used for various applications, such as attack detection
and attack type recognition, where inferring attack details is possible.
Through the lens of the ARC feature (reflecting the network's gradient behavior),
we also obtain insights on why
networks are vulnerable and why adversarial training works well as a defense.
Although our method is specific to PGD-like attacks due to strong assumptions, it is
\textbf{(1)} intuitive (human-interpretable due to simplicity and not creating a deep model);
\textbf{(2)} light-weighted (requires no auxiliary deep model);
\textbf{(3)} non-intrusive (requires no change to the network architecture or weights);
\textbf{(4)} data-undemanding (can generalize with only a few samples).


\section{Adversarial Response Characteristics}
\label{sec:2}

A network $\vf(\cdot)$ maps the input 
$\vx\in\mathbb{R}^M$ into a pre-softmax output $\vy\in\mathbb{R}^N$, where the
maximum element after softmax corresponds to the class prediction $\hat{c}(\vx)$,
which should match with the ground truth $c(\vx)$.
Then, a typical adversarial attack~\citep{bim,madry} aims to find an imperceptible
adversarial perturbation
$\vr \in \mathbb{R}^M$ that induces misclassification, \emph{i.e.}, $\argmax_n
f_n(\vx+\vr) \neq c(\vx)$ where
$\|\vr\|_p \leq \varepsilon$,
$\vx+\vr\in [0,1]^M$,
and $f_n(\cdot)$ is the $n$-th element of vector
function $\vf(\cdot)$.

\begin{figure}[t]
    \includegraphics[width=1.0\linewidth]{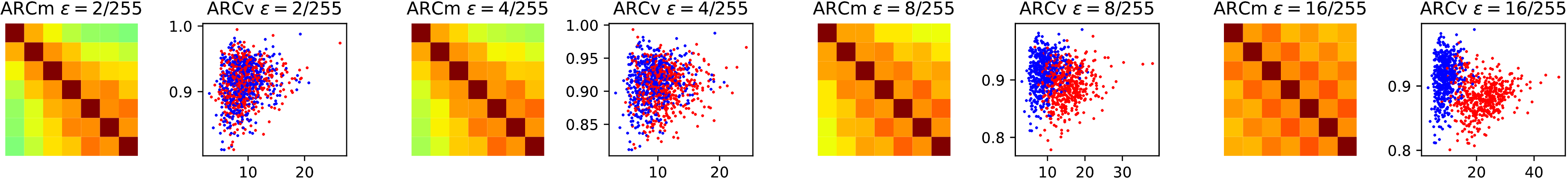} \newline
    \includegraphics[width=1.0\linewidth]{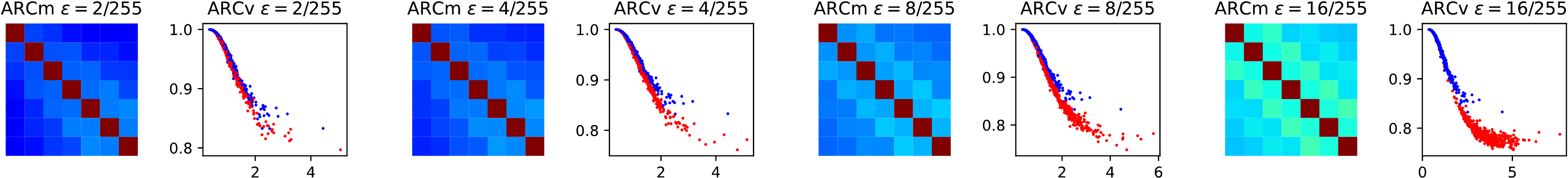} \newline
    \includegraphics[width=1.0\linewidth]{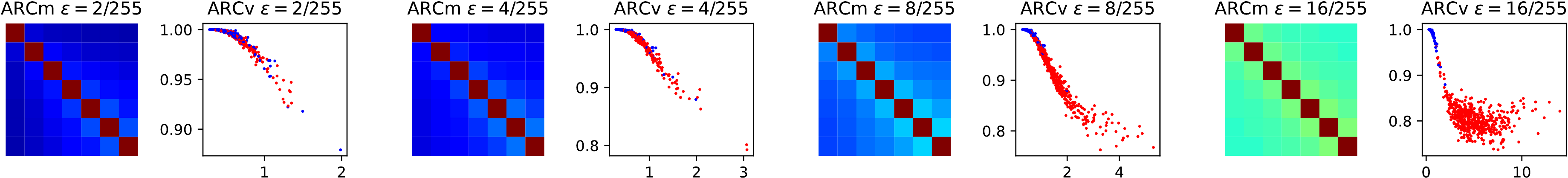}
    \caption{The ARC features (\emph{i.e.} ARC matrix/vector)
	of adversarial examples created by the BIM attack.
	1\textsuperscript{st} row: ResNet-18 on CIFAR-10;
	2\textsuperscript{nd} row: ResNet-152 on ImageNet;
	3\textsuperscript{rd} row: SwinT-B-IN1K on ImageNet.
	Blue and red dots in the scatter plots correspond to the benign and
	adversarial examples, respectively.
	The cluster centers of the ARC vector correlate with the perturbation magnitude $\varepsilon$.
    }
    \label{fig:arcfeat}
\end{figure}

According to \citep{fgsm}, a neural network is vulnerable as the ``locally
linear'' property is triggered by attack.
Thus, we assume that the network $\vf(\cdot)$ behaves
relatively non-linear against benign examples, while 
relatively linear against adversarial examples.
Then, $\vf(\cdot)$ can be approximated by the first-order Taylor
expansion around an either benign or adversarial sample $\tilde{\vx}$:
\begin{equation}
\tilde{\vx}\triangleq \vx+\vr, \quad
f_n(\tilde{\vx}+\bm\delta) \approx f_n(\tilde{\vx}) + \bm\delta^T \nabla
	f_n(\tilde{\vx}),\quad \forall n \in \{1,2,\ldots,N\},
\end{equation}
where $\bm\delta$ is a small vector exploiting the local area around
the point $\tilde{\vx}$,
and the gradient vector $\nabla f_n(\cdot)$ is the $n$-th row of
the Jacobian $\nabla \vf(\cdot)$ of size $N{\times}M$.
We name the twice-perturbed $\tilde{\vx}+\bm\delta$ as ``exploitation
vector''.
This equation means in order to reflect linear behavior,
the first-order gradient $\nabla f_n(\cdot)$
is expected to remain in high consistency (or similarity)  in the local area regardless
of $\bm\delta$.
In contrast, when the input $\tilde{\vx}$ is not adversarial ($\vr=\bm{0}$), neither Taylor
approximation nor the gradient consistency is expected to hold.
Next, the gradient consistency will be quantized to verify our conjecture,
and reveal the difference between benign and adversarial inputs.

\textbf{Adversarial Response Characteristics (ARC).}
Using random noise as $\bm\delta$ does not lead to a stable pattern of change in a series of
exploitation vectors $\{\tilde{\vx}+\bm\delta_t\}_{t=0,1,\ldots,T}$.
Instead, we use Basic Iterative Method (BIM)~\citep{bim}
to make $\vf(\cdot)$ more linear starting from $\tilde{\vx}$, which
means to ``continue'' the attack if $\tilde{\vx}$ is already adversarial,
or ``restart'' otherwise.
However, the ground-truth label for an arbitrary $\tilde{\vx}$ is \emph{unknown}.
Since PGD-like attacks tend to make the ground-truth least-likely based on
our observation, we treat the least-likely prediction $\check{c}(\vx)$ as the label.
Then, the BIM iteratively maximizes the cross entropy loss
$L_\text{CE}(\tilde{\vx}+\bm\delta, \check{c}(\vx))$ via projected gradient ascent as
\begin{equation}
\bm\delta_{t+1} \leftarrow \text{Clip}_\Omega\Big( \bm\delta_t + \alpha
	\sign[\nabla L_\text{CE}(\tilde{\vx}+\bm\delta_t, \check{c}(\vx))]\Big),
	\quad t=0,1,2,\ldots,T,
\end{equation}
where  $\text{Clip}_\Omega(\cdot)$ clips the perturbation to the $L_p$ bound
centered at $\tilde{\vx}$, and $\bm\delta_0 = \bm{0}$.
If the input $\tilde{\vx}$ is benign, then the network behavior is expected to
change from ``very non-linear`` to ``somewhat-linear'' during the process; 
if the input $\tilde{\vx}$ is already adversarially perturbed,
then the process will ``continue'' the attack, making the model 
even more ``linear'' -- we call this \emph{Sequel Attack Effect} (SAE).

To quantize the extent of ``linearity'', we measure the model's gradient
consistency across exploitation vectors with cosine similarity.
For each $f_n(\cdot)$, we construct a
matrix $\mS_n$ of shape $(T{+}1,T{+}1)$:
\begin{equation}
	s^{(i,j)}_n=\cos\big[ \nabla f_n(\tilde{\vx}+\bm\delta_i), \nabla
	f_n(\tilde{\vx}+\bm\delta_j) \big],
	\quad \forall i,j=0,1,\ldots,T.
\end{equation}
As the model $\vf(\cdot)$ becomes more ``linear'' to the input
(higher gradient consistency),
the off-diagonal values in $\mS_n$ are expected to gradually increase from
the top-left to the bottom-right corner.
Note that the attack may not necessarily make all $f_n(\cdot)$ behave linear,
so we select the most representative cosine matrix
with the highest mean as the \emph{ARC matrix}:
$\mS_*\triangleq \mS_{n^*}$, where $n^*=\arg\max_{n} \sum_{i,j} s^{(i,j)}_{n}$.

Due to the resemblance of the ARC matrix to the Laplacian function
with the matrix diagonal being the center,
we simplify it into a two-dimensional \emph{ARC vector}
$(A,\sigma)$ by fitting $\mathcal{L}(i,j;
A,\sigma)=A\exp(-|i-j|/\sigma)$ with Levenberg-Marquardt algorithm~\citep{scipy},
where $i,j$ are matrix row and column indexes,
while $A$ and $\sigma$ are function parameters.
For brevity, we abbreviate the ARC matrix as ``ARCm'', and the ARC vector as ``ARCv''.
The overall process is summarized in Fig.~\ref{fig:diag}.

\begin{figure}[t]
	\centering

	\begin{subfigure}[t]{0.32\linewidth}
		\includegraphics[width=\linewidth]{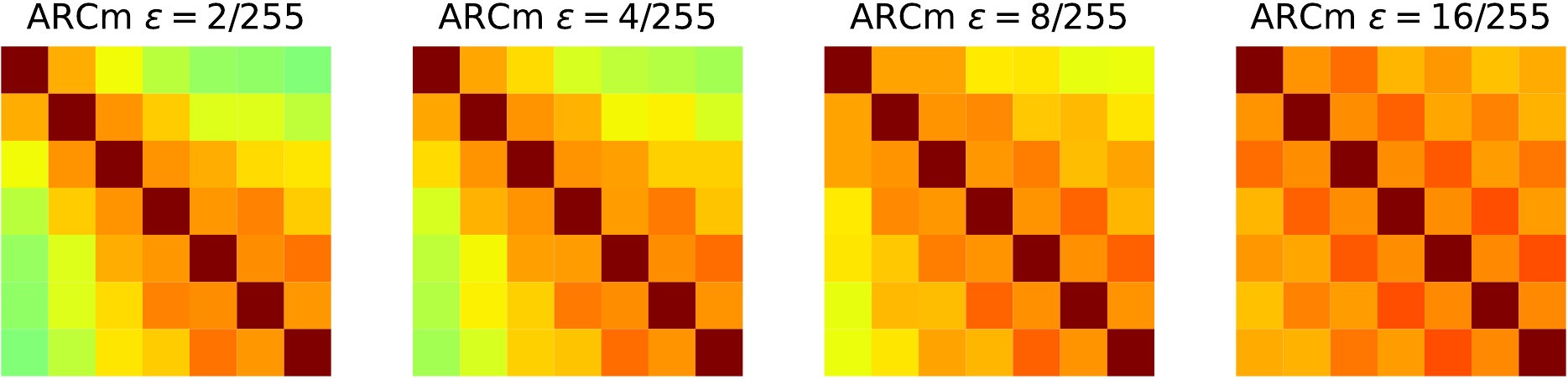}\\
		\includegraphics[width=\linewidth]{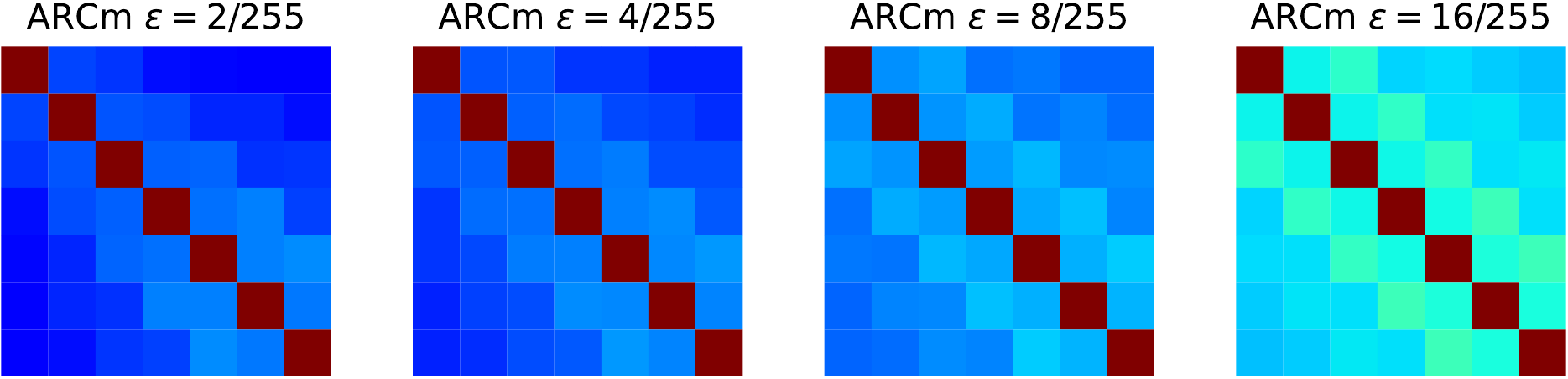}\\
		\includegraphics[width=\linewidth]{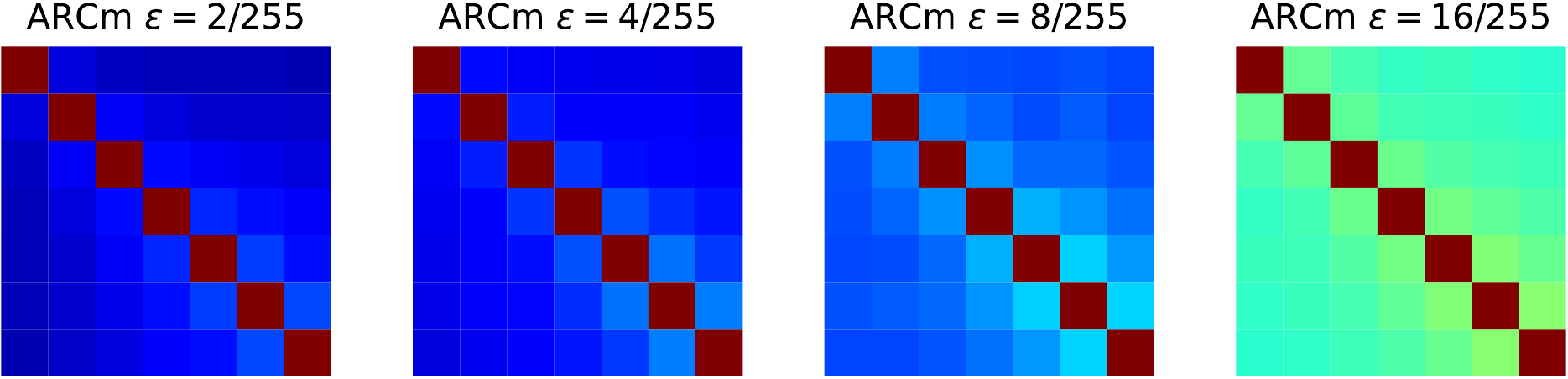}
	\end{subfigure}
	\hfill
	\unskip \vrule\
%
	\begin{subfigure}[t]{0.32\linewidth}
		\includegraphics[width=\linewidth]{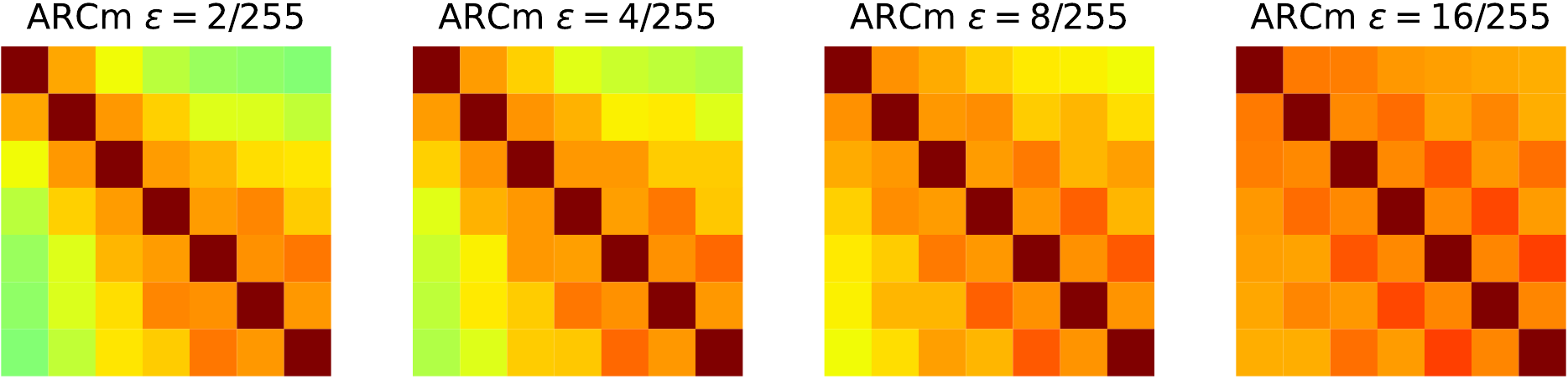}\\
		\includegraphics[width=\linewidth]{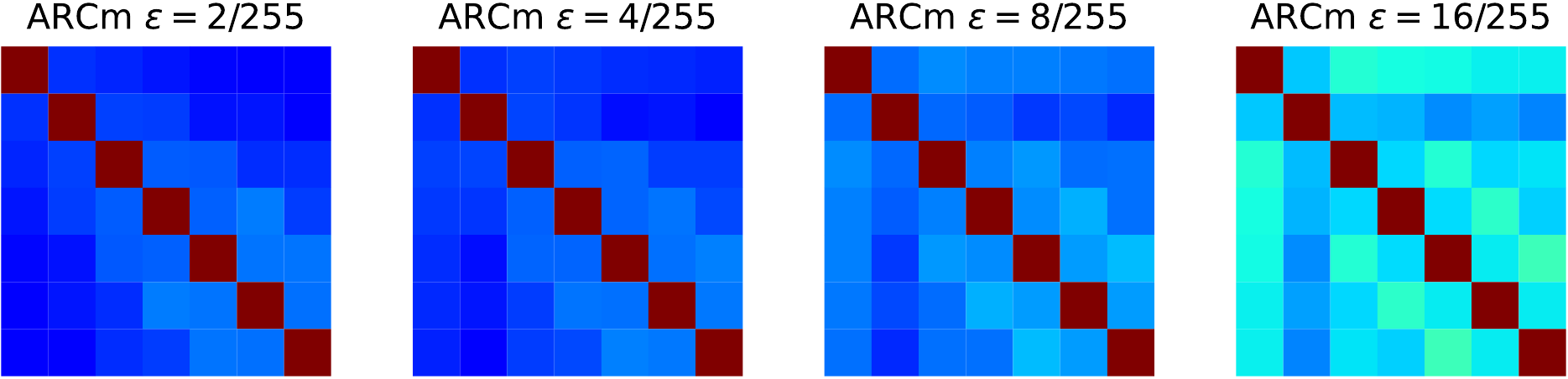}\\
		\includegraphics[width=\linewidth]{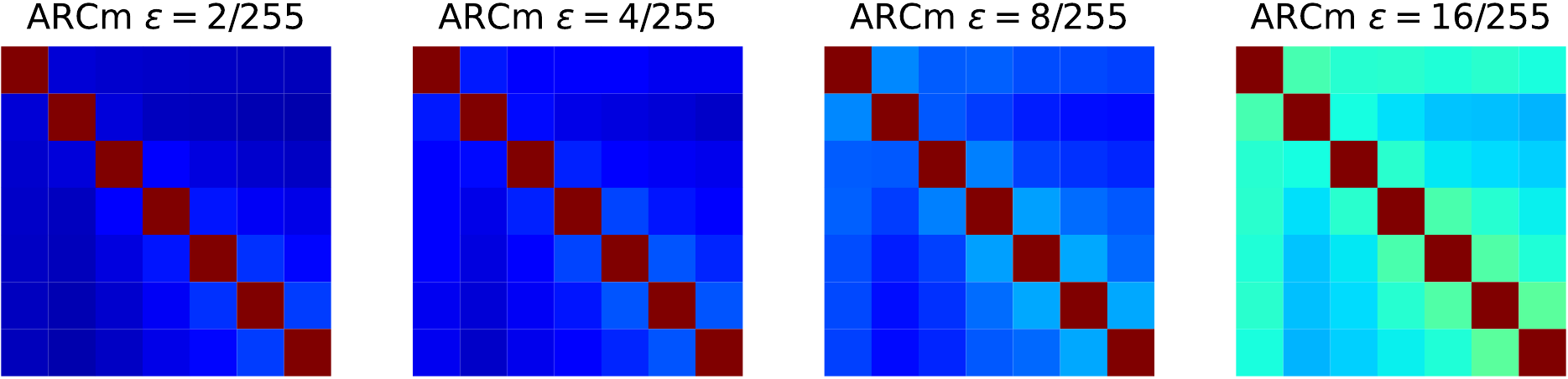}
	\end{subfigure}
	\hfill
	\unskip \vrule\
%
	\begin{subfigure}[t]{0.32\linewidth}
		\includegraphics[width=\linewidth]{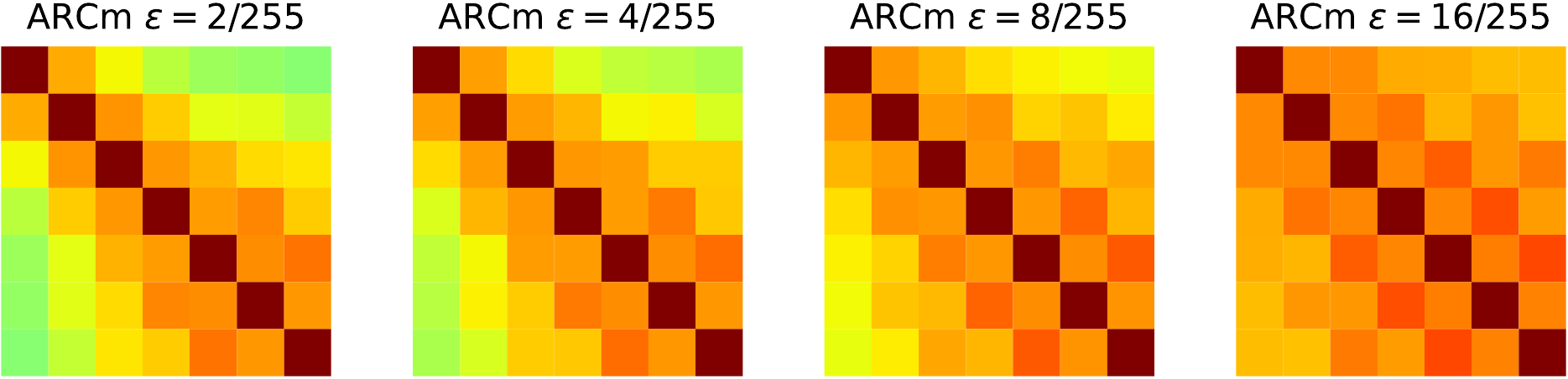}\\
		\includegraphics[width=\linewidth]{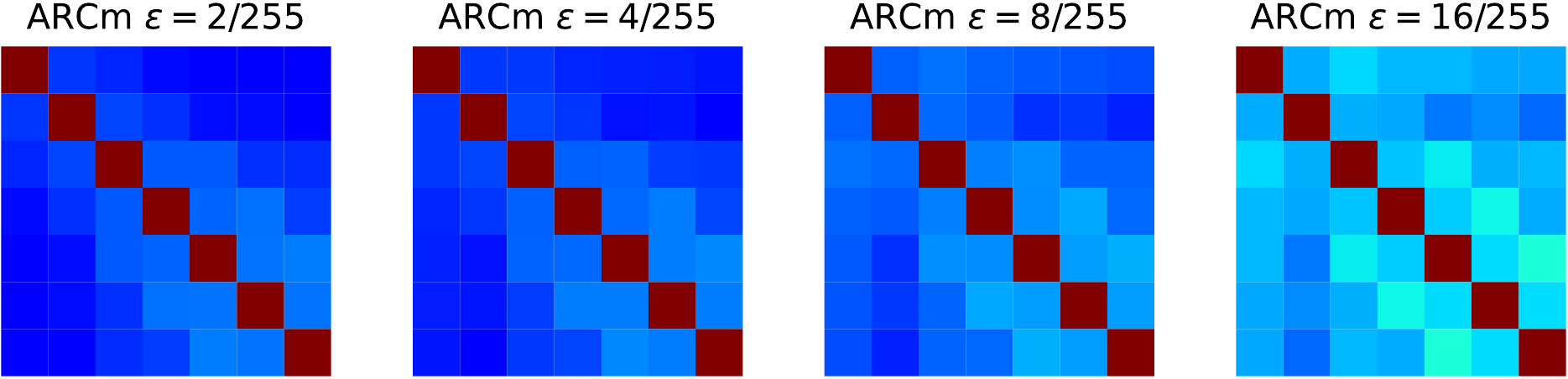}\\
		\includegraphics[width=\linewidth]{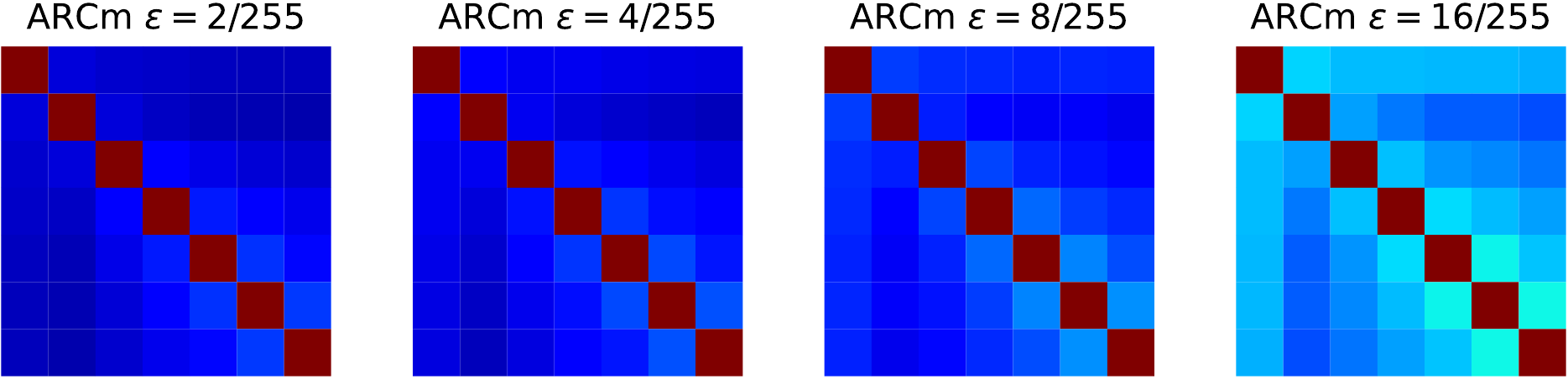}
	\end{subfigure}

	\caption{ARCm with adversarial examples created by PGD
	(left), MIM (middle), and APGD (right) attacks.
	The three rows correspond to ResNet-18, ResNet-152, and SwinT-B-IN1K,
	respectively.
	It is clear that PGD-like attacks qualitatively manifest similar SAE through ARCm.
	}
	\label{fig:pgdmimapgd}
\end{figure}

\textbf{Visualizing Sequel Attack Effect (SAE).}
We compute ARCm based on some benign examples using $T{=}48$, as shown in
Fig.~\ref{fig:diag}.
The trend of being gradually ``linear'' (higher cosine similarity) along the
diagonal is found across architectures.
Thus, SAE is similar to ``continuing'' an attack from halfway on the diagonal in
such a large ARCm.
As illustrated in Fig.~\ref{fig:arcfeat}, already adversarially perturbed
input (using BIM) leads to larger cosine similarity at the very first exploitation vectors
as perturbation magnitude $\varepsilon$ increases from $0$ to $16/255$.
Meanwhile, the cluster separation for ARCv is more and more clear.
Thus, a clear and gradually changing pattern can be seen in ARCm and ARCv.
This pattern is even valid and clear for the state-of-the-art ImageNet models.
In brief, SAE is reflected by higher gradient consistency in ARCm,
or greater $\sigma$ and smaller $A$ in ARCv.
Similar results for other PGD-like attacks in Fig.~\ref{fig:pgdmimapgd}
indicate the possibility of generalization among them
with only training samples from the BIM attack. 
We adopt SVM afterward to retain interpretability and simplicity.

\textbf{Uniqueness of SAE to PGD-Like Attack.}
Whether SAE can be consistently triggered depends on the following conditions
simultaneously being \emph{true}:
\textbf{(I)} whether the input is adversarially perturbed by an \emph{iterative} projected
gradient update method for many steps;
\textbf{(II)} whether the attack leverages the \emph{first-order gradient} of the model;
\textbf{(III)} whether the $L_p$ boundary types are the same for the two stages, \emph{i.e.},
attack and exploitation vectors;
\textbf{(IV)} whether the loss functions for the two stages are the same;
\textbf{(V)} whether the labels used (if any) for the two stages are relevant.
Namely, only when the attack and exploitation vectors
``match'', can SAE be uniquely triggered as the exploitation
vectors ``continue'' an attack, or they will ``restart'' an attack.
Thus, in Fig.~\ref{fig:diag},
Fig.~\ref{fig:arcfeat} and Fig.~\ref{fig:pgdmimapgd}, all the conditions are true
as they involve PGD-like attacks.
Due to the strong assumptions, the SAE being insensitive to non-PGD-like attacks (\emph{e.g.},
\citet{cw}) is a \emph{limitation}.
However, the unique SAE meanwhile shows a possibility of inferring the attack
details leveraging the above conditions.
SAE is the trace of PGD-like attacks.
Ablations for these five conditions are presented in Sec.~\ref{sec:5}.

\textbf{Adaptive Attack.}
Adaptive attacks can be designed against defenses~\citep{adaptive},
detection~\citep{carlini}.
To avoid SAE in ARCm, an adaptive attack must reach a point
where the corresponding ARCm has a mean value as small as that for
benign examples.
Intuitively, an adaptive attack has to simultaneously solve
$\min_\vr \| \mS_*(\vx+\vr) \|_F$ (Frobenius norm) alongside its original attack goal.
It however requires the gradient of Jacobians, namely at least $T+1$ Hessian
matrices, \emph{i.e.}, $\nabla^2 f_n(\cdot)$ of size $M\times M$
to perform gradient descent.
This is computationally prohibitive as in the typical ImageNet setting
(\emph{i.e.}, $M{=}3{\times}224{\times}224$), a Hessian in \verb|float32| precision
needs $84.4$GiB memory.
At this point, the cost of adaptive attack that hides SAE is much higher than
computing ARC.
Instead, a viable way to avoid SAE is to use non-PGD-like attacks
that break the SAE uniqueness conditions.
This paper focuses on characterizing the unique trace of existing
PGD-like attacks, instead of a general detection or defense method.


\section{Applications of ARC Feature}
\label{sec:3}

In order to quantitatively support the effectiveness of ARC/SAE, we adopt it for two
potential tasks, namely attack detection and attack type recognition.
Attack detection aims to identify the attempt to adversarially perturb an image
\emph{even if} it fails to change the prediction (but leaves a trace).\footnote{
In practice, it is undesirable to wait and react until the attack has succeeded.}
Attack type recognition aims to identify whether an adversarial example is created
by PGD-like attacks.
Our method relies on the uniqueness of SAE to PGD-like attacks for the two tasks.

\textbf{Informed Attack Detection}
determines whether an arbitrary input $\tilde{\vx}$ is adversarially perturbed,
while the perturbation magnitude $\varepsilon$ is \emph{known}.
It can be viewed as a binary classification problem, where the input is ARCv of $\tilde{\vx}$,
and the output $1$ indicates ``adversarially perturbed'', while $0$ indicates ``unperturbed''.
Thus, for a given $\varepsilon=2^k/255$ where $k\in\{1,2,3,4\}$, a
corresponding SVM~\citep{sklearn} classifier
$h_k(\tilde{\vx})\in\{0,1\}$
can be trained using some benign ($\varepsilon{=}0$)
samples and their adversarial counterparts ($\varepsilon{=}2^k{/}255$).
Even if the training data only involves the BIM attack,
we expect generalization for other PGD-like attacks from visualization results despite domain shift.

\textbf{Uninformed Attack Detection}
determines whether an arbitrary input $\tilde{\vx}$ is adversarially
perturbed, while the perturbation magnitude $\varepsilon$ is \emph{unknown}.
It can be viewed as an ordinal regression~\citep{orcnn} problem, where the input
is ARCv, and the output is the estimation of $k$, namely $\hat{k}\in\{0,1,2,3,4\}$.
The corresponding estimate of $\varepsilon$ is $\hat{\varepsilon}
=\bm{1}\{\hat{k}>0\}2^{\hat{k}}/255$, where $\bm{1}\{\cdot\}$ is the indicator function.
Specifically, this is implemented as a series of binary classifiers (SVM), where the
$k$-th ($k{\neq}0$) classifier predicts whether the level of perturbation is greater or equal
to $k$, \emph{i.e.}, whether $\hat{k}\geqslant k$.
Note, based on our visualization, the ARCv cluster of adversarial
examples is moving away from that of benign examples as $\varepsilon$
(or $k$) increases.
This means the ARCv of an adversarial example with $\hat{k}\geqslant k$
will also cross the decision boundary of the $k$-th SVM $h_k(\cdot)$.
Namely the SVM $h_k(\cdot)$ can also tell whether $\hat{k}\geqslant k$,
and thus can be reused.
Finally, the ordinal regression model can be expressed as the sum of prediction over
the SVMs: $\hat{k}= \sum_{k\in\{1,2,3,4\}} h_k(\tilde{\vx})$.
A perturbation is detected as long as $\hat{k}>0$.
Estimating $k$ (or $\varepsilon$) for $\tilde{\vx}$ is similar to matching
its ARCm position inside a much larger ARCm calculated starting from a benign
example.
But, the estimate does not have to be
precise, because the detection is already successful once any of the SVMs
correctly raises an alert.

Although a detector in practice knows completely nothing about a potential attack
including the attack type, evaluation of uninformed attack detection
with \emph{known} attack type is enough.
Regarding the performance for uninformed attack detection given a specific attack
type as a conditional performance, the expected performance in the wild
can be calculated as the sum of conditional performance weighted
by the prior probabilities that the corresponding attack happens.

\textbf{Inferring Attack Details.}
Due to the SAE uniqueness in Sec.~\ref{sec:2},
once the attack is detected, we can also predict that the attack:
(I) performs projected gradient update iteratively;
(II) uses the first-order gradient of $\vf(\cdot)$;
(III) uses the same type of $L_p$ bound as exploitation vectors ($L_\infty$ by default);
(IV) uses the same loss as exploitation vectors ($L_\text{CE}(\cdots)$ by default);
(V) uses a ground-truth label which is relevant to the least-likely
class $\check{c}(\tilde{\vx})$ used for exploitation vectors
(in many cases $\check{c}(\tilde{\vx})$ is exactly the ground-truth).
In other words, model prediction can be corrected
into the least-likely class $\check{c}(\tilde{\vx})$ upon detection.
Namely, the disadvantage of ARC being insensitive to non-PGD-like attacks
is meanwhile advantage of being able to infer attack details of PGD-like attacks.

\textbf{Attack Type Recognition} determines whether an adversarial input
is created by PGD-like attacks in the uninformed setting for forensics purposes.
The corresponding binary classifier can be built upon the previously discussed
detectors, because SAE only responds to PGD-like attacks.


\begin{figure}[t]
	\centering
	\small

	\renewcommand{\thesubfigure}{f\arabic{subfigure}}
	\captionsetup[subfigure]{justification=centering,font=tiny}

	\begin{subfigure}[t]{0.16\linewidth}
		\includegraphics[width=\linewidth]{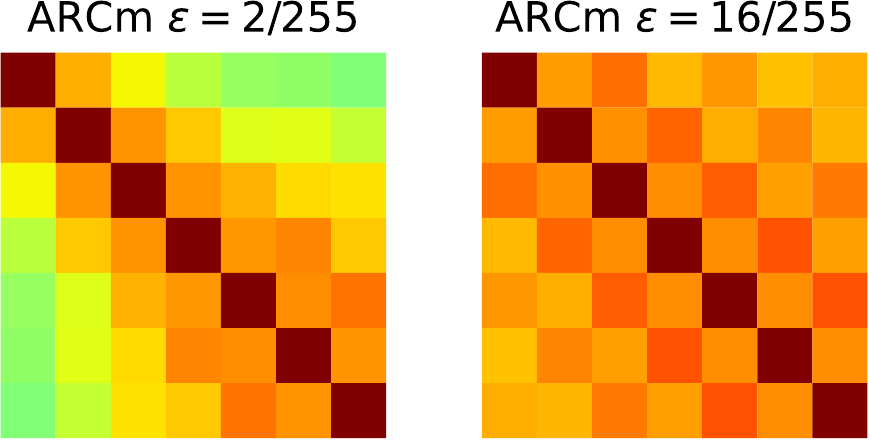}
		\vspace{-1.7em}
		\caption{BIM[]; ARC[]}
	\end{subfigure}
	\hfill
	\begin{subfigure}[t]{0.16\linewidth}
		\includegraphics[width=\linewidth]{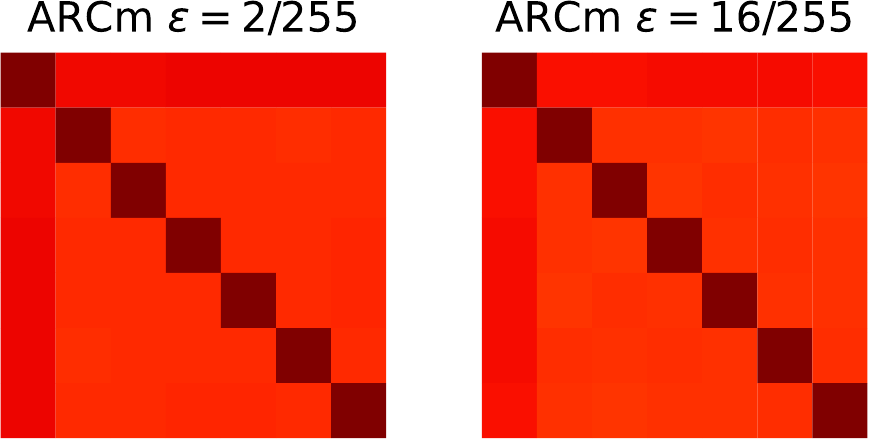}
		\vspace{-1.7em}
		\caption{BIM[]; Gaussian}
	\end{subfigure}
	\hfill
	\begin{subfigure}[t]{0.16\linewidth}
		\includegraphics[width=\linewidth]{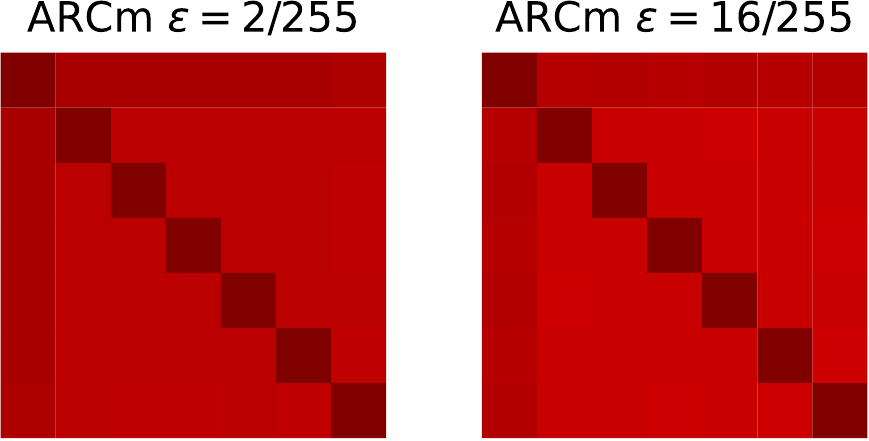}
		\vspace{-1.7em}
		\caption{BIM[]; Uniform}
	\end{subfigure}
	\hfill
	\begin{subfigure}[t]{0.16\linewidth}
		\includegraphics[width=\linewidth]{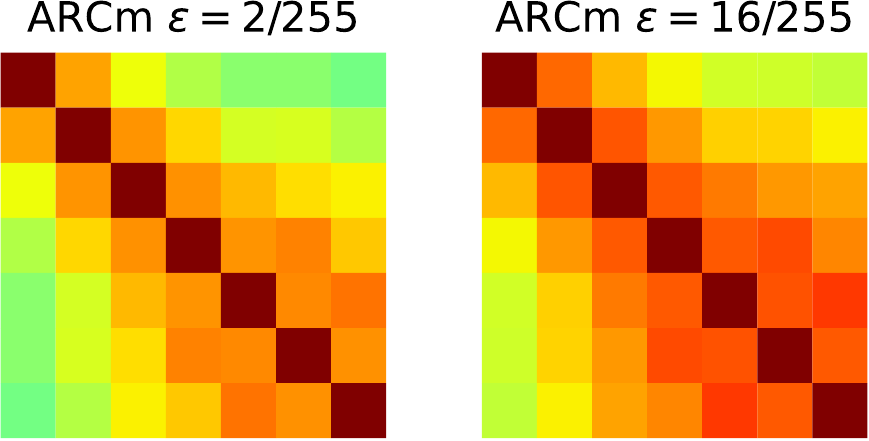}
		\vspace{-1.7em}
		\caption{FGSM[\xmark Iter]; ARC[]}
	\end{subfigure}
	\hfill
	\begin{subfigure}[t]{0.16\linewidth}
		\includegraphics[width=\linewidth]{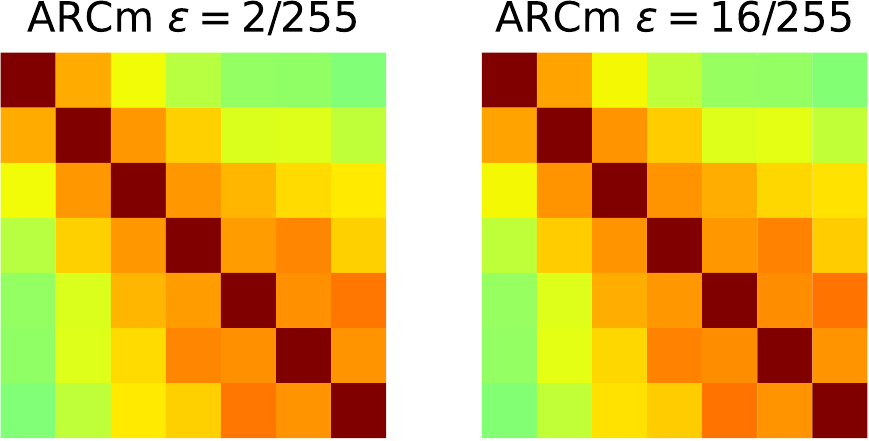}
		\vspace{-1.7em}
		\caption{Gaussian; ARC[]}
	\end{subfigure}
	
	\begin{subfigure}[t]{0.16\linewidth}
		\includegraphics[width=\linewidth]{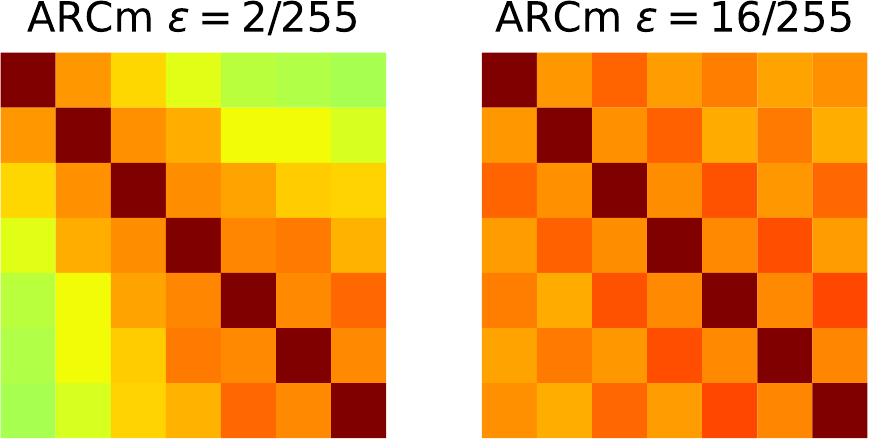}
		\vspace{-1.7em}
		\caption{BIM[$L_2$]; ARC[]}
	\end{subfigure}
	\hfill
	\begin{subfigure}[t]{0.16\linewidth}
		\includegraphics[width=\linewidth]{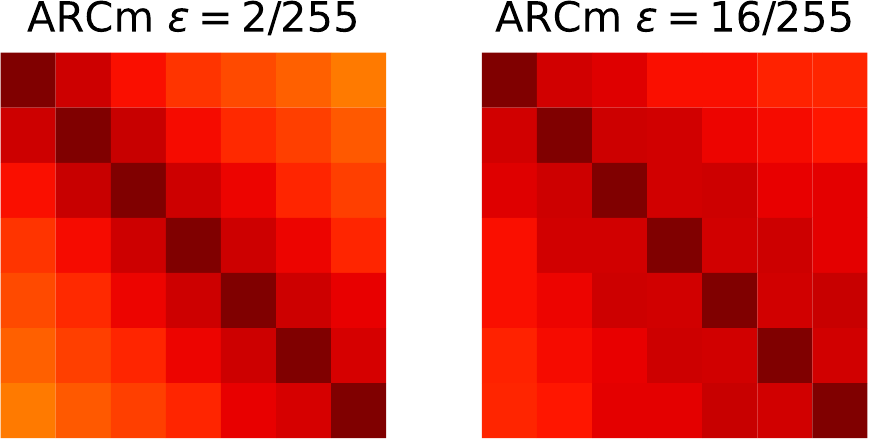}
		\vspace{-1.7em}
		\caption{BIM[]; ARC[$L_2$]}
	\end{subfigure}
	\hfill
	\begin{subfigure}[t]{0.16\linewidth}
		\includegraphics[width=\linewidth]{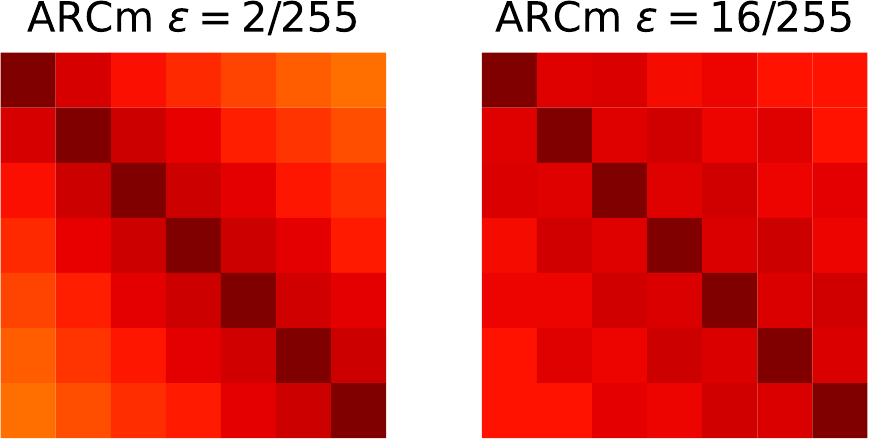}
		\vspace{-1.7em}
		\caption{BIM[$L_2$]; ARC[$L_2$]}
	\end{subfigure}
	\hfill
	\begin{subfigure}[t]{0.16\linewidth}
		\includegraphics[width=\linewidth]{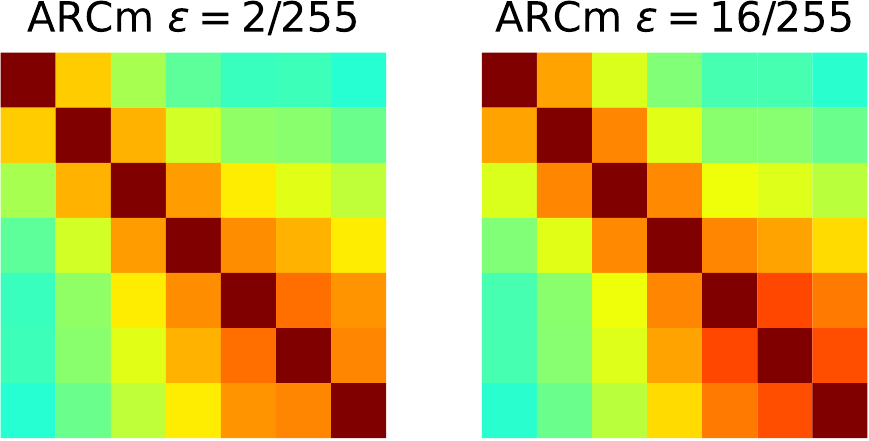}
		\vspace{-1.7em}
		\caption{BIM[]; ARC[$\hat{c}(\vx)$]}
	\end{subfigure}
	\hfill
	\begin{subfigure}[t]{0.16\linewidth}
		\includegraphics[width=\linewidth]{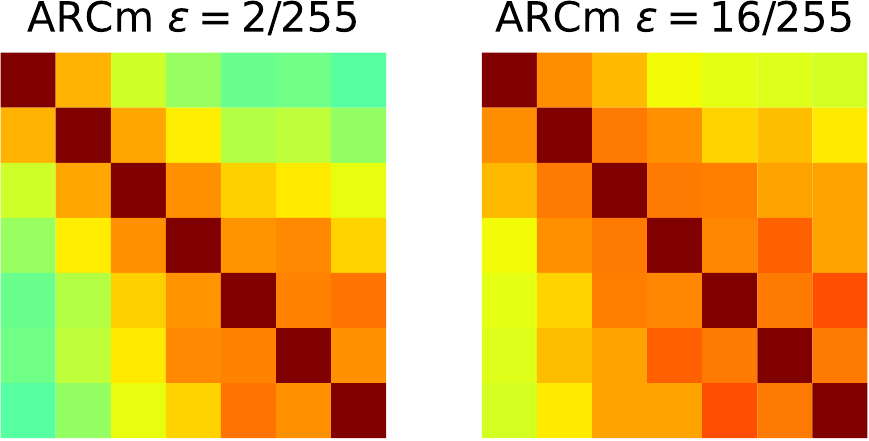}
		\vspace{-1.7em}
		\caption{BIM[]; ARC[$c?$]}
	\end{subfigure}

	\begin{subfigure}[t]{0.16\linewidth}
		\includegraphics[width=\linewidth]{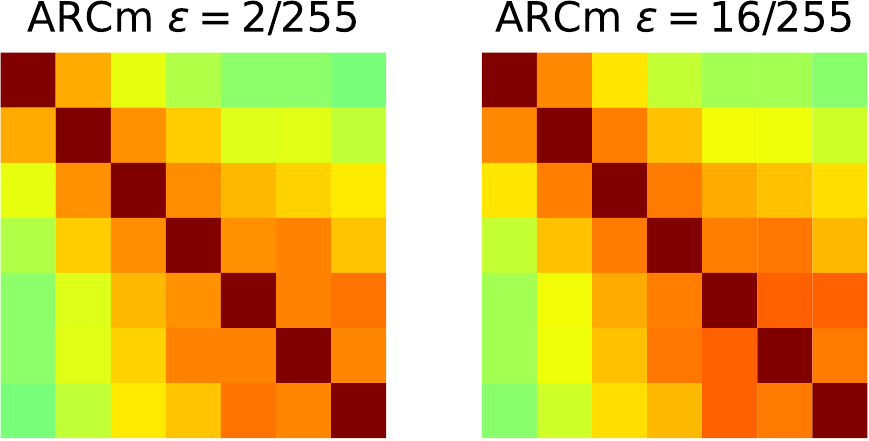}
		\vspace{-1.7em}
		\caption{BIM[DLR]; ARC[]}
	\end{subfigure}
	\hfill
	\begin{subfigure}[t]{0.16\linewidth}
		\includegraphics[width=\linewidth]{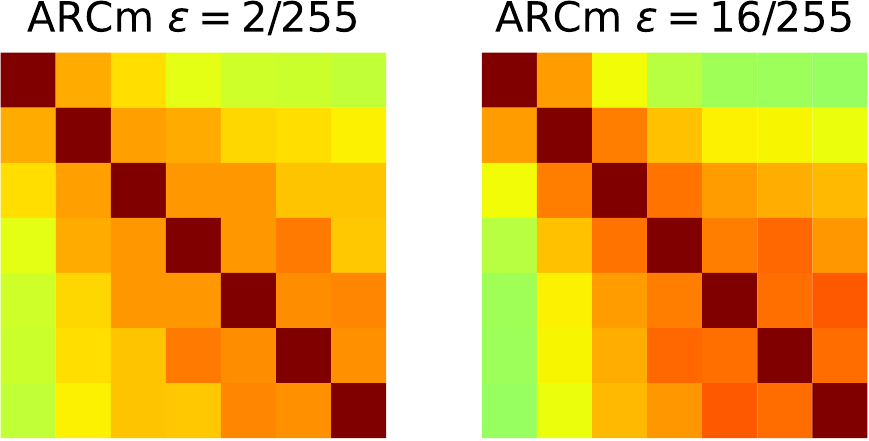}
		\vspace{-1.7em}
		\caption{BIM[]; ARC[DLR]}
	\end{subfigure}
	\hfill
	\begin{subfigure}[t]{0.16\linewidth}
		\includegraphics[width=\linewidth]{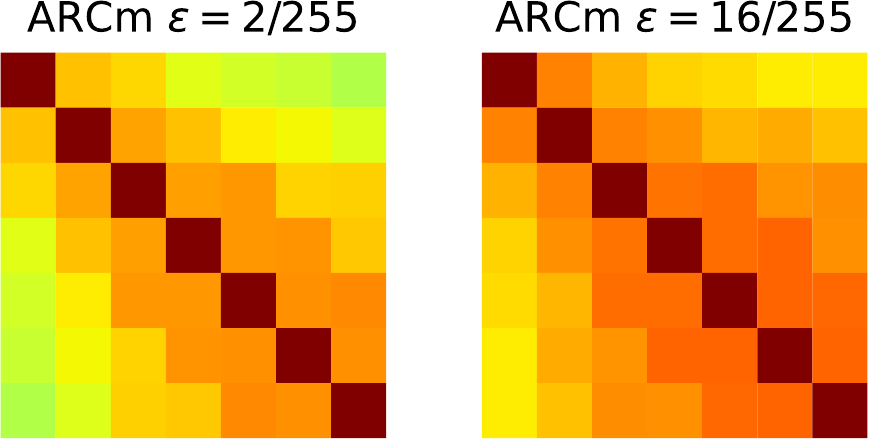}
		\vspace{-1.7em}
		\caption{\scalebox{0.8}{BIM[DLR]; ARC[DLR]}}
	\end{subfigure}
	\hfill
	\begin{subfigure}[t]{0.16\linewidth}
		\includegraphics[width=\linewidth]{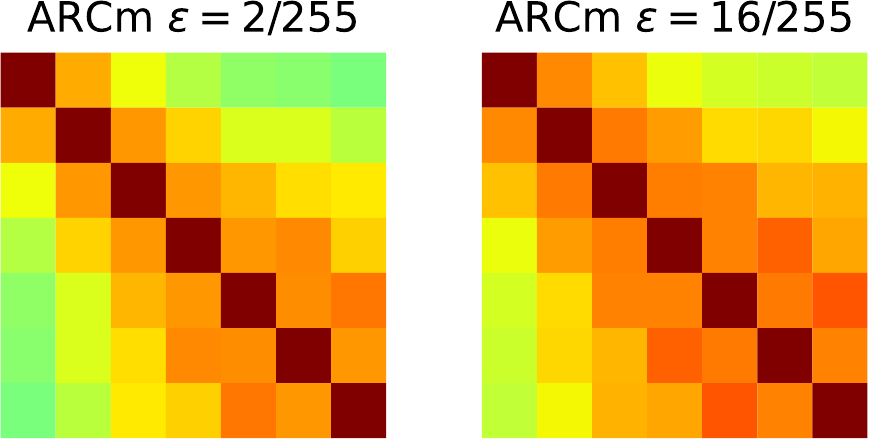}
		\vspace{-1.7em}
		\caption{NES[\xmark $\nabla\vf$]; ARC[]}
	\end{subfigure}
	\hfill
	\begin{subfigure}[t]{0.16\linewidth}
		\includegraphics[width=\linewidth]{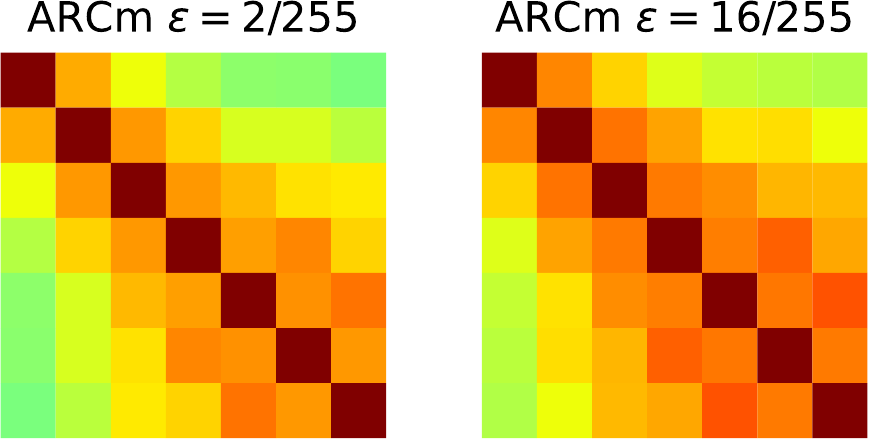}
		\vspace{-1.7em}
		\caption{\scalebox{0.8}{SPSA[\xmark $\nabla\vf$]; ARC[]}}
	\end{subfigure}

	\caption{Ablation of SAE uniqueness by adjusting exploitation vectors for ARC.
	Each subfigure of ARCm pair has two annotations:
	(1) attack and its settings, where empty brackets mean default setting
	unless overridden:
	[$L_p$ is $L_\infty$; Loss is $L_\text{CE}$; \cmark (is) iterative;
	\cmark (can access) gradient $\nabla \vf(\cdot)$];
	(2) exploitation vector settings, \emph{e.g.} ``ARC[]''
	with the default setting [$L_p$ is $L_\infty$; Loss is $L_\text{CE}$;
	Label is $\check{c}(\cdot)$]. The ``$c?$'' means random guess.
	This figure is supplementary to Tab.~\ref{tab:nonpgdlike}.
	}
	\label{fig:fig4}
\end{figure}

\section{Experiments}
\label{sec:4}

In this section, we quantitatively verify the effectiveness of the ARC features
in two applications, and the effectiveness of the post-processing step
under an \emph{extremely limited setting}.
Unlike related works, the MNIST evaluation is omitted, as
the corresponding conclusions may not hold~\citep{carlini} on CIFAR-10, let alone
ImageNet.
We evaluate ResNet-18~\citep{resnet} on CIFAR-10~\citep{cifar};
ResNet-152~\citep{resnet} and SwinT-B-IN1K~\citep{swint}
on ImageNet~\citep{imagenet} with their official pre-trained weights
(advantage of being non-intrusive).
Our code is implemented based on PyTorch~\citep{pytorch}.

\textbf{ARC Feature Parameter.}
For the BIM attack for exploitation vectors, we set step number $T=6$,
and step size $\alpha=\nicefrac{2}{255}$ under the $L_\infty$ bound with $\varepsilon=
\nicefrac{8}{255}$.
Note, the mean value of ARCm will tend to $1$ with a larger $T$, making
ARCv less separatable.
We choose $T=6$ to clearly visualize the value changes within ARCm, but
this does not necessarily lead to the best performance.

\textbf{Training.}
We train SVMs $h_k(\cdot)$ with RBF kernel. We randomly select $\bm{50}$ training
samples from CIFAR-10, and perturb them using \emph{only} BIM with
magnitude $\varepsilon=\nicefrac{2}{255}, \nicefrac{4}{255}, \nicefrac{8}{255},
\nicefrac{16}{255}$, respectively.
Then each of the four $h_k(\cdot)$ is trained with ARCv of the benign
($\varepsilon=0$) samples and perturbed ($\varepsilon=2^k/255$) samples.
Likewise, for ImageNet we randomly select $\bm{50}$ training samples and train SVM
in a similar setting separately for ResNet-152 and SwinT-B-IN1K.
The weight for benign examples can be adjusted for training to control the
False Positive Rate (FPR).

\textbf{Testing.}
For CIFAR-10, all $10000$ testing data and their perturbed versions with
different $\varepsilon$ are used to test our SVM.
For ImageNet, we randomly choose $512$ testing samples
due to costly Jacobian computation.
A wide range of adversarial attacks are involved, including
(1) PGD-like attacks: BIM~\citep{bim}, PGD~\citep{madry},
MIM~\citep{mim}, APGD~\citep{autoattack}, AutoAttack (AA)~\citep{autoattack};
(2) Non-PGD-like attacks:
(2.1) other white-box attacks: FGSM~\citep{fgsm}, C\&W~\citep{cw} (we use
$\varepsilon\in \{0.5, 1.0, 2.0, 3.0\}$ in $L_2$ case), FAB~\citep{fab}, FMN~\citep{fmn};
(2.2) transferability attacks: DI-FGSM~\citep{difgsm}, TI-FGSM~\citep{tifgsm}
(using ResNet-50 as proxy);
(2.3) score-based black-box methods: NES~\citep{nes}, SPSA~\citep{spsa}, Square~\citep{square}.
%
%
AutoAttack is regarded as PGD-like because APGD is its most significant component
for success rate.
Details can be found in the supplementary code.

\textbf{Metrics.}
The SVMs are evaluated with Detection Rate (DR, \emph{a.k.a.}, True
Positive Rate) and False Positive Rate (FPR).
For inferring the ground-truth label, we report the original accuracy
for perturbed examples (denoted as ``Acc'') and that after correction
(denoted as ``Acc*'').
Mean Average Error (MAE) is also reported for ordinal regression.
Accuracy is reported for attack type recognition.

\subsection{Application of ARC: Attack Detection}
\label{sec:41}


\begin{table}
	\caption{Informed and Uninformed (the ``$\varepsilon{=}?$''
column) Attack Detection. All numbers are percentages with the ``\%''
sign omitted, except for MAE. Numbers greater than 50\% are in bold font.}
	\label{tab:pgd-like}

\resizebox{\linewidth}{!}{
\setlength{\tabcolsep}{3pt}

\begin{tabular}{cc|cccc|cccc|cccc|cccc|ccccc}

	\toprule

\multirow{2}{*}{\textbf{\thead{Dataset\\Model}}} & \multirow{2}{*}{\textbf{Attack}} & \multicolumn{4}{c|}{$\epsilon=2/255$} & \multicolumn{4}{c|}{$\epsilon=4/255$} & \multicolumn{4}{c|}{$\epsilon=8/255$} & \multicolumn{4}{c|}{$\epsilon=16/255$} & \multicolumn{5}{c}{$\epsilon=?$}\tabularnewline
\cline{3-23} \cline{4-23} \cline{5-23} \cline{6-23} \cline{7-23} \cline{8-23} \cline{9-23} \cline{10-23} \cline{11-23} \cline{12-23} \cline{13-23} \cline{14-23} \cline{15-23} \cline{16-23} \cline{17-23} \cline{18-23} \cline{19-23} \cline{20-23} \cline{21-23} \cline{22-23} \cline{23-23}
 &  & DR & FPR & Acc & Acc{*} & DR & FPR & Acc & Acc{*} & DR & FPR & Acc & Acc{*} & DR & FPR & Acc & Acc{*} & MAE & DR & FPR & Acc & Acc{*}\tabularnewline

	\midrule

\multirow{5}{*}{\makecell{CIFAR-10\\ResNet-18}} & BIM & 0.0 & 0.0 & 33.5 & 33.5 & 0.0 & 0.0 & 6.4 & 6.4 & 32.3 & 1.5 & 0.4 & 17.8 & \textbf{79.2} & 1.1 & 0.0 & \textbf{62.4} & 1.55 & 30.9 & 1.5 & 10.1 & 30.7\tabularnewline
 & PGD & 0.0 & 0.0 & 33.7 & 33.7 & 0.0 & 0.0 & 6.4 & 6.4 & 33.0 & 1.5 & 0.4 & 18.6 & \textbf{81.2} & 1.1 & 0.0 & \textbf{64.8} & 1.54 & 31.5 & 1.5 & 10.1 & 31.5\tabularnewline
 & MIM & 0.0 & 0.0 & 30.4 & 30.4 & 0.0 & 0.0 & 6.5 & 6.5 & 37.5 & 1.5 & 0.4 & 22.3 & \textbf{84.5} & 1.1 & 0.0 & \textbf{67.4} & 1.50 & 33.6 & 1.5 & 9.3 & 32.4\tabularnewline
 & APGD & 0.0 & 0.0 & 29.3 & 29.3 & 0.0 & 0.0 & 5.1 & 5.1 & 36.9 & 1.5 & 0.2 & 20.7 & \textbf{78.8} & 1.1 & 0.0 & \textbf{55.8} & 1.53 & 31.5 & 1.5 & 8.7 & 28.0\tabularnewline
 & AA & 0.0 & 0.0 & 27.4 & 27.4 & 0.0 & 0.0 & 2.1 & 2.1 & 37.3 & 1.5 & 0.0 & 20.6 & \textbf{78.4} & 1.1 & 0.0 & \textbf{55.6} & 1.53 & 31.6 & 1.5 & 7.4 & 26.8\tabularnewline
\rowcolor{black!10} & \textbf{?} & 0.0 & 0.0 & 30.9 & 30.9 & 0.0 & 0.0 & 5.3 & 5.3 & 35.4 & 1.5 & 0.3 & 20.0 & \textbf{80.4} & 1.1 & 0.0 & \textbf{61.2} & 1.53 & 31.8 & 1.5 & 9.1 & 29.9\tabularnewline

\midrule

\multirow{5}{*}{\makecell{ImageNet\\ResNet-152}} & BIM & 0.0 & 0.0 & 0.0 & 0.0 & 4.7 & 1.4 & 0.0 & 0.0 & 20.5 & 1.4 & 0.0 & 0.0 & \textbf{91.6} & 1.4 & 0.0 & 0.4 & 1.36 & 30.6 & 1.6 & 0.0 & 0.1\tabularnewline
 & PGD & 0.0 & 0.0 & 0.0 & 0.0 & 4.7 & 1.4 & 0.0 & 0.0 & 18.8 & 1.4 & 0.0 & 0.0 & \textbf{85.9} & 1.4 & 0.0 & 0.0 & 1.44 & 28.9 & 1.6 & 0.0 & 0.0\tabularnewline
 & MIM & 0.0 & 0.0 & 0.0 & 0.0 & 2.3 & 1.4 & 0.0 & 0.0 & 4.7 & 1.4 & 0.0 & 0.0 & \textbf{81.2} & 1.4 & 0.0 & 0.0 & 1.52 & 23.8 & 1.6 & 0.0 & 0.2\tabularnewline
 & APGD & 0.0 & 0.0 & 0.0 & 0.0 & 2.0 & 1.4 & 0.0 & 0.0 & 11.3 & 1.4 & 0.0 & 0.0 & \textbf{61.7} & 1.4 & 0.0 & 0.4 & 1.59 & 19.7 & 1.6 & 0.0 & 0.1\tabularnewline
 & AA & 0.0 & 0.0 & 0.0 & 0.0 & 2.5 & 1.4 & 0.0 & 0.0 & 10.7 & 1.4 & 0.0 & 0.0 & \textbf{61.5} & 1.4 & 0.0 & 0.0 & 1.59 & 19.9 & 1.6 & 0.0 & 0.0\tabularnewline
\rowcolor{black!10} & \textbf{?} & 0.0 & 0.0 & 0.0 & 0.0 & 3.2 & 1.4 & 0.0 & 0.0 & 13.2 & 1.4 & 0.0 & 0.0 & \textbf{76.3} & 1.4 & 0.0 & 0.2 & 1.50 & 24.6 & 1.6 & 0.0 & 0.1\tabularnewline

\midrule

\multirow{5}{*}{\makecell{ImageNet\\SwinT-B-IN1K}} & BIM & 4.1 & 1.6 & 6.1 & 6.2 & 13.7 & 2.0 & 0.0 & 8.4 & \textbf{77.3} & 2.0 & 0.0 & \textbf{74.0} & \textbf{97.9} & 0.2 & 0.0 & \textbf{97.9} & 0.96 & 49.1 & 2.0 & 1.5 & 47.3\tabularnewline
 & PGD & 3.9 & 1.6 & 2.3 & 3.1 & 16.4 & 2.0 & 0.0 & 10.9 & \textbf{72.7} & 2.0 & 0.0 & \textbf{68.8} & \textbf{98.4} & 0.2 & 0.0 & \textbf{98.4} & 1.01 & 48.6 & 2.0 & 0.6 & 45.9\tabularnewline
 & MIM & 1.6 & 1.6 & 0.0 & 1.6 & 10.2 & 2.0 & 0.0 & 10.2 & \textbf{63.3} & 2.0 & 0.0 & \textbf{63.3} & \textbf{93.8} & 0.2 & 0.0 & \textbf{93.8} & 1.09 & 43.8 & 2.0 & 0.0 & 43.8\tabularnewline
 & APGD & 1.4 & 1.6 & 0.0 & 1.0 & 5.3 & 2.0 & 0.0 & 4.5 & 32.6 & 2.0 & 0.0 & 25.2 & \textbf{65.0} & 0.2 & 0.0 & \textbf{51.0} & 1.37 & 29.4 & 2.0 & 0.0 & 23.2\tabularnewline
 & AA & 1.8 & 1.6 & 0.0 & 1.0 & 5.7 & 2.0 & 0.0 & 4.3 & 31.6 & 2.0 & 0.0 & 25.0 & \textbf{68.4} & 0.2 & 0.0 & \textbf{54.1} & 1.37 & 29.5 & 2.0 & 0.0 & 23.2\tabularnewline
\rowcolor{black!10} & \textbf{?} & 2.6 & 1.6 & 1.7 & 2.6 & 10.2 & 2.0 & 0.0 & 7.7 & \textbf{55.5} & 2.0 & 0.0 & \textbf{51.2} & \textbf{84.7} & 0.2 & 0.0 & \textbf{79.0} & 1.16 & 40.1 & 2.0 & 0.4 & 36.7\tabularnewline

\bottomrule
\end{tabular}}

\label{tab:pgdlike}
\end{table}

\begin{wrapfigure}[9]{t}{0.5\linewidth}
	\includegraphics[width=0.32\linewidth]{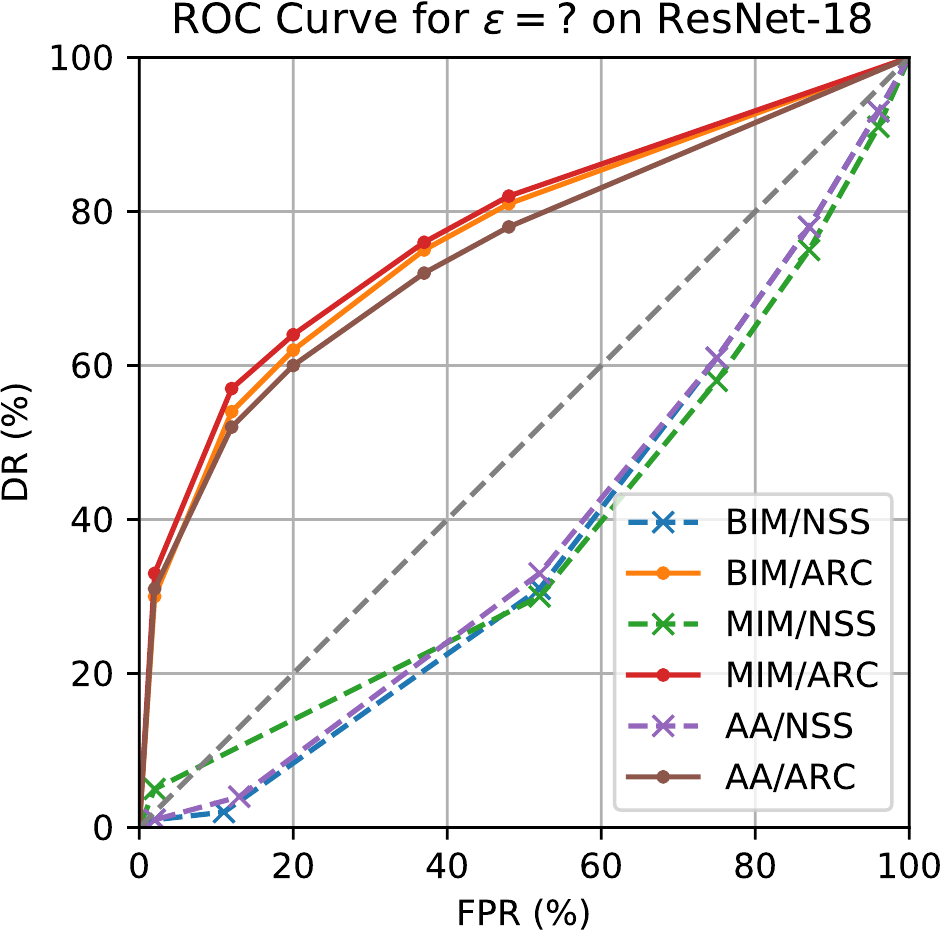}
	\includegraphics[width=0.32\linewidth]{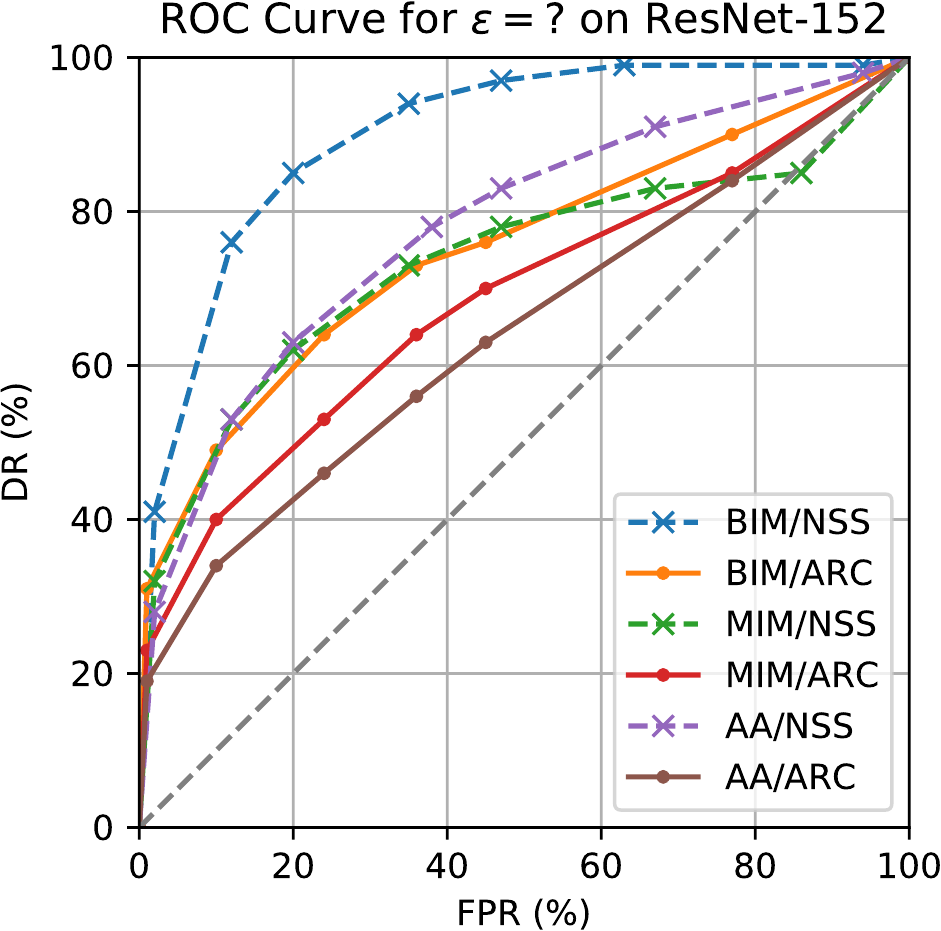}
	\includegraphics[width=0.32\linewidth]{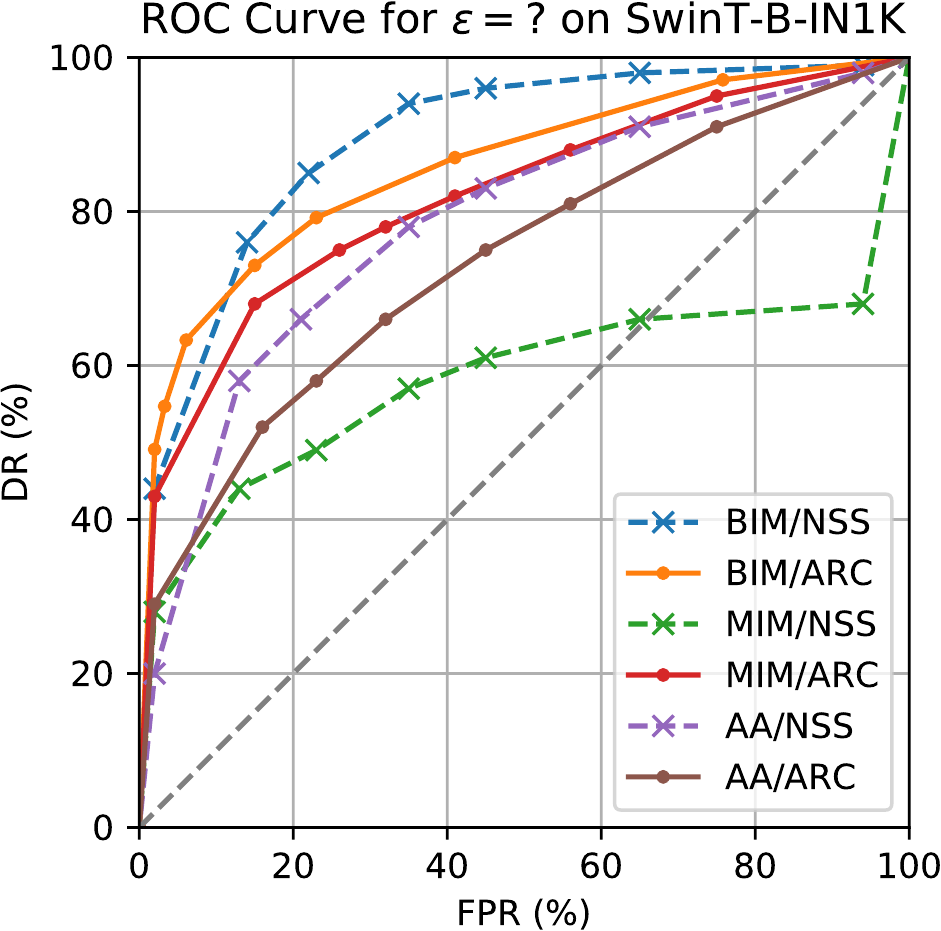}
	\caption{ROC of SVMs in Tab.~\ref{tab:pgdlike} \& Tab.~\ref{tab:sota}.}
	\label{fig:roc}
\end{wrapfigure}

For each network, the corresponding SVMs are trained and
evaluated as shown in Tab.~\ref{tab:pgdlike}.
Columns with a concrete $\varepsilon$ value are informed attack detection,
while the ``$\varepsilon{=}?$`` column is uninformed attack detection.
As can be expected from visualization results, the ARCv clusters are gradually
becoming separatable with $\varepsilon$ increasing, hence the increase
of DR.
Notably, the large perturbations (\emph{i.e.}, $\varepsilon=16/255$) are
hard to defend~\citep{LLS}, but can be consistently detected
across architectures.
The ARC feature is especially effective for Swin-Transformer, because this model
transitions faster from being non-linear to being linear than other architectures.
Such characteristics are beneficial for SAE.

Upon detection of an attack, our method can correct the prediction into the
least-likely class as a post-processing step.
Its success rate depends on whether the attack is efficient to make
the ground-truth class least-likely, and whether the network is easy for the
attack to make a class least-likely.
From Tab.~\ref{tab:pgdlike}, both ResNet-18 and SwinTransformer
have such a property and lead to high classification accuracy after correction.
For ResNet-152, the least-likely label is merely relevant (not identical) to the ground truth
due to network property during attack, hence leading to effective detection
but not correction (this will be explained in the next subsection).
In contrast, the correction method performs best on
Swin-Transformer, as it can restore classification accuracy from $0.4\%$
to $36.7\%$ even if both the concrete type of PGD-like attack and $\varepsilon$ are unknown
(``Attack=?'' row and ``$\varepsilon{=}?$'' column in Tab.~\ref{tab:pgdlike}),
assuming flat prior.
By adjusting the weights assigned to benign examples, the decision boundary
of SVMs can be moved and hence influence the FPR, as shown
in Fig.~\ref{fig:roc}.

\subsection{Sequel Attack Effect as Unique Trace of PGD-like Attacks}
\label{sec:42}


\begin{table}
\caption{Ablation of SAE uniqueness by varying attacks.
	The row (t1) is regarded as a baseline, and the notation ``..''
	means ``same as the baseline'' in order to ease comparison.
	SAE will only show consistent effectiveness across architectures
	when the conditions in Sec.~\ref{sec:2} are satisfied.
	}

\resizebox{\linewidth}{!}{
\setlength{\tabcolsep}{3pt}

\begin{tabular}{cc|cccc|ccc|ccccc|ccccc|ccccc}

	\toprule

\multirow{2}{*}{\#} & \multicolumn{5}{c|}{\textbf{Attack}} & \multicolumn{3}{c|}{\textbf{ARC}} & \multicolumn{5}{c|}{ResNet-18 w/ $\epsilon=?$} & \multicolumn{5}{c|}{ResNet-152 w/ $\epsilon=?$} & \multicolumn{5}{c}{SwinT-B-IN1K w/ $\epsilon=?$}\tabularnewline
\cline{2-24} \cline{3-24} \cline{4-24} \cline{5-24} \cline{6-24} \cline{7-24} \cline{8-24} \cline{9-24} \cline{10-24} \cline{11-24} \cline{12-24} \cline{13-24} \cline{14-24} \cline{15-24} \cline{16-24} \cline{17-24} \cline{18-24} \cline{19-24} \cline{20-24} \cline{21-24} \cline{22-24} \cline{23-24} \cline{24-24} 
 & \textbf{Name} & $L_{p}$ & Loss & Iter. & $\nabla\bm{f}(\cdot)$ & $L_{p}$ & Loss & Label & MAE & DR & FPR & Acc & Acc{*} & MAE & DR & FPR & Acc & Acc{*} & MAE & DR & FPR & Acc & Acc{*}\tabularnewline

	\midrule

\rowcolor{black!10}t1 & BIM & $\infty$ & CE & Yes & Yes & $\infty$ & CE & $\check{c}(\bm{x})$ & 1.55 & 30.9 & 1.5 & 10.1 & 30.7 & 1.36 & 30.6 & 1.6 & 0.0 & 0.1 & 0.96 & 49.1 & 2.0 & 1.5 & 47.3\tabularnewline

\hline

t2 & BIM & 2 & .. & .. & .. & .. & .. & .. & 1.27 & 49.9 & 1.5 & 2.6 & 39.0 & 1.98 & 3.5 & 1.6 & 0.2 & 0.2 & 2.02 & 1.0 & 2.0 & 1.4 & 1.8\tabularnewline

\rowcolor{black!10}t3 & BIM & .. & DLR & .. & .. & .. & .. & .. & 1.98 & 2.1 & 1.5 & 10.5 & 10.6 & 1.63 & 18.9 & 1.6 & 0.0 & 0.6 & 1.44 & 27.5 & 2.0 & 1.8 & 6.6\tabularnewline

t4 & FGSM & .. & .. & No & .. & .. & .. & .. & 1.96 & 3.4 & 1.5 & 30.3 & 29.5 & 1.63 & 18.6 & 1.6 & 8.4 & 6.8 & 1.44 & 27.1 & 2.0 & 44.9 & 32.4\tabularnewline

\rowcolor{black!10}t5 & C\&W & $2$ & C\&W & .. & .. & .. & .. & .. & 1.99 & 1.2 & 1.5 & 0.0 & 0.0 & 2.02 & 2.3 & 1.6 & 0.0 & 0.0 & 2.03 & 1.6 & 2.0 & 0.0 & 0.0\tabularnewline

t6 & FAB & .. & FAB & .. & .. & .. & .. & .. & 1.99 & 1.0 & 1.5 & 10.6 & 10.5 & 2.00 & 2.5 & 1.6 & 9.2 & 9.2 & 2.03 & 0.8 & 2.0 & 9.4 & 9.4\tabularnewline

\rowcolor{black!10}t7 & FMN & .. & FMN & .. & .. & .. & .. & .. & 1.99 & 1.4 & 1.5 & 8.8 & 8.6 & 2.02 & 2.1 & 1.6 & 0.0 & 0.0 & 2.03 & 0.8 & 2.0 & 0.0 & 0.0\tabularnewline

t8 & DI-FGSM & .. & DI-FGSM & .. & No & .. & .. & .. & 1.98 & 2.2 & 1.5 & 42.9 & 42.0 & 1.98 & 3.5 & 1.6 & 27.9 & 27.5 & 1.87 & 8.2 & 2.0 & 67.2 & 62.1\tabularnewline

\rowcolor{black!10}t9 & TI-FGSM & .. & TI-FGSM & .. & No & .. & .. & .. & 1.98 & 1.9 & 1.5 & 59.4 & 58.3 & 2.00 & 2.9 & 1.6 & 40.0 & 39.1 & 2.02 & 1.6 & 2.0 & 72.3 & 70.9\tabularnewline

t10 & NES & .. & .. & .. & No & .. & .. & .. & 1.94 & 4.7 & 1.5 & 38.6 & 39.4 & 1.98 & 3.1 & 1.6 & 28.3 & 27.3 & 2.02 & 1.6 & 2.0 & 50.6 & 49.4\tabularnewline

\rowcolor{black!10}t11 & SPSA & .. & .. & .. & No & .. & .. & .. & 1.97 & 3.0 & 1.5 & 39.2 & 39.1 & 2.00 & 3.1 & 1.6 & 29.9 & 28.9 & 2.00 & 2.7 & 2.0 & 52.7 & 50.6\tabularnewline

t12 & Square & .. & Square & .. & No & .. & .. & .. & 1.99 & 1.6 & 1.5 & 85.7 & 84.3 & 2.02 & 2.1 & 1.6 & 68.6 & 67.4 & 1.84 & 10.2 & 2.0 & 77.9 & 70.1\tabularnewline

\rowcolor{black!10}t13 & Gaussian & .. & N/A & No & No & .. & .. & .. & 1.99 & 1.7 & 1.5 & 87.0 & 85.6 & 2.00 & 2.7 & 1.6 & 75.2 & 73.2 & 2.00 & 3.1 & 2.0 & 82.4 & 79.7\tabularnewline

t14 & Uniform & .. & N/A & No & No & .. & .. & .. & 1.99 & 1.8 & 1.5 & 86.6 & 85.0 & 1.97 & 4.1 & 1.6 & 73.6 & 70.9 & 1.84 & 10.2 & 2.0 & 81.8 & 73.2\tabularnewline

\bottomrule

\end{tabular}}

	\label{tab:nonpgdlike}
\end{table}

The SAE is unique to PGD-like attacks, as it requires five
conditions listed in Sec.~\ref{sec:2} to hold for consistent effectiveness.
To clarify this, we change the attack settings
(quantitatively in Tab.~\ref{tab:nonpgdlike}), or the exploitation
vector for ARCm (qualitatively on CIFAR10 in Fig.~\ref{fig:fig4}),
and then review these conditions:
\begin{enumerate}[label=(\Roman*).,leftmargin=*,nosep,noitemsep]
	\item \underline{Iterative attack (Iter.).}
The single-step version of PGD, \emph{i.e.}, FGSM (t4, f4) does not effectively
exploit the search space within the $L_p$ bound, and hence will not easily
trigger linearity and SAE.
Only Swin Transformer slightly reacts against FGSM due to its own characteristics
of being easy to be turned linear.
Thus, SAE requires the attack to be iterative;

\item \underline{Gradient access} ($\nabla\vf(\cdot)$).
Transferability-based attacks (t8, t9) use proxy model gradients to
create adversarial examples, and hence could not trigger SAE.
NES (t10, f14) and SPSA (t11, f15) can be seen as PGD using gradients
estimated from only network logits, but can still not trigger SAE as
it cannot efficiently trigger linearity.
Neither does Square attack (t12).
Thus, SAE requires that the attacks use the target model gradient;

\item \underline{Same} $L_p$ \underline{bound.}
When the attack is BIM in $L_2$ bound (t2, f6), SAE will no longer be triggered
for ImageNet models, because the change of $L_p$ influences
the perturbation search process.
However, SAE is still triggered for CIFAR-10 possibly due to relatively
low-dimensional search space.
This means CIFAR-10 property does not necessarily generalize to ImageNet.
When ARC has been changed accordingly (f7, f8), the feature
clusters are still separatable.
Thus, SAE requires the same type of $L_p$ bound for consistent effectiveness;

\item \underline{Same loss.}
When the loss for BIM is switched from $L_\text{CE}$ to
DLR~\citep{autoattack} (t3, f11), the SAE is significantly reduced.
However, if exploitation vectors are also created using DLR (f12, f13),
SAE will be triggered again.
Thus, SAE requires a consistent loss;

\item \underline{Relevant label.}
When the most-likely label $\hat{c}(\tilde{\vx})$ is used for exploitation
vectors, it leads to the least significant SAE (f9).
Besides, even a random label ($c?$) leads to moderate SAE (f10), while
the least-likely label $\check{c}(\tilde{\vx})$ (which is ground-truth
label in many cases) leads to distinct SAE (f1).
The most significant SAE correspond to $\check{c}(\tilde{\vx})=c(\vx)$.
This means that to maximize cross-entropy, the local linearity of a large portion
of output functions $f_n(\cdot)$ has been triggered.
Thus, SAE requires a relevant label (if any) for exploitation vectors.

\end{enumerate}

When the exploitation vectors are created using random noise (f2, f3), SAE
is not triggered.
Neither does random noise as an attack trigger SAE (t13, t14, f5).
Other non-PGD-like attacks (t5, t6, t7) do not trigger SAE either.
A special case is targeted PGD-like attack, where the creation of exploitation
vectors needs to use negative cross-entropy loss on the most-likely
label to reach a similar level of effectiveness (this
paper focuses on the default untargeted attack to avoid complication).

The non-PGD attacks, or PGD variants do not meet all conditions cannot
consistently trigger SAE across architectures because they provide a
less ``matching'' starting point for exploitation vectors, and hence make
the BIM for exploitation vectors ``restart'' an attack, where the
network behaves non-linear again.
Only when all the conditions are satisfied will SAE be consistently
triggered across different architectures, especially for ImageNet models.
As for label correction,
PGD-like attacks can effectively leak the ground-truth labels
in the adversarial example, as long as the network allows the attack to
easily reduce the corresponding logit value to the lowest.


\subsection{Comparison with Previous Attack Detection Methods}
\label{sec:43}


\begin{table}
\caption{Comparison with existing methods that are compatible with our problem setting.}

\resizebox{\linewidth}{!}{
\setlength{\tabcolsep}{2pt}%
\begin{tabular}{c|c|ccccc|ccccc|ccccc|ccccc|ccccc}

	\toprule

\multirow{2}{*}{\textbf{Method}} & \multirow{2}{*}{\textbf{Metric}} & \multicolumn{5}{c|}{\textbf{BIM}} & \multicolumn{5}{c|}{\textbf{PGD}} & \multicolumn{5}{c|}{\textbf{MIM}} & \multicolumn{5}{c|}{\textbf{APGD}} & \multicolumn{5}{c}{\textbf{AA}}\tabularnewline
\cline{3-27} \cline{4-27} \cline{5-27} \cline{6-27} \cline{7-27} \cline{8-27} \cline{9-27} \cline{10-27} \cline{11-27} \cline{12-27} \cline{13-27} \cline{14-27} \cline{15-27} \cline{16-27} \cline{17-27} \cline{18-27} \cline{19-27} \cline{20-27} \cline{21-27} \cline{22-27} \cline{23-27} \cline{24-27} \cline{25-27} \cline{26-27} \cline{27-27}
 &  & \multicolumn{1}{c|}{2/255} & \multicolumn{1}{c|}{4/255} & \multicolumn{1}{c|}{8/255} & \multicolumn{1}{c|}{16/255} & ? & \multicolumn{1}{c|}{2/255} & \multicolumn{1}{c|}{4/255} & \multicolumn{1}{c|}{8/255} & \multicolumn{1}{c|}{16/255} & ? & \multicolumn{1}{c|}{2/255} & \multicolumn{1}{c|}{4/255} & \multicolumn{1}{c|}{8/255} & \multicolumn{1}{c|}{16/255} & ? & \multicolumn{1}{c|}{2/255} & \multicolumn{1}{c|}{4/255} & \multicolumn{1}{c|}{8/255} & \multicolumn{1}{c|}{16/255} & ? & \multicolumn{1}{c|}{2/255} & \multicolumn{1}{c|}{4/255} & \multicolumn{1}{c|}{8/255} & \multicolumn{1}{c|}{16/255} & ?\tabularnewline

\midrule

\multicolumn{27}{c}{\textbf{CIFAR10 ResNet-18}}\tabularnewline

	\rowcolor{black!10}\cellcolor{white}\multirow{2}{*}{NSS~\citep{nss}} & DR & 0.0 & 0.0 & 0.0 & 0.1 & 0.5 & 0.0 & 0.0 & 0.0 & 0.1 & 0.5 & 0.0 & 0.0 & 0.0 & 0.1 & 4.7 & 0.0 & 0.0 & 0.3 & 0.2 & 0.8 & 0.0 & 0.0 & 0.3 & 0.2 & 0.8\tabularnewline
 & FPR & 0.0 & 0.0 & 1.8 & 1.5 & 2.5 & 0.0 & 0.0 & 1.8 & 1.5 & 2.5 & 0.0 & 0.0 & 1.8 & 1.5 & 2.5 & 0.0 & 0.0 & 1.8 & 1.5 & 2.5 & 0.0 & 0.0 & 1.8 & 1.5 & 2.5\tabularnewline

\rowcolor{black!10}\cellcolor{white}\multirow{2}{*}{ARC} & DR & 0.0 & 0.0 & 32.3 & \textbf{79.2} & 30.9 & 0.0 & 0.0 & 33.0 & \textbf{81.2} & 31.5 & 0.0 & 0.0 & 37.5 & \textbf{84.5} & 33.6 & 0.0 & 0.0 & 36.9 & \textbf{78.8} & 31.5 & 0.0 & 0.0 & 37.3 & \textbf{78.4} & 31.6\tabularnewline
 & FPR & 0.0 & 0.0 & 1.5 & 1.1 & 1.5 & 0.0 & 0.0 & 1.5 & 1.1 & 1.5 & 0.0 & 0.0 & 1.5 & 1.1 & 1.5 & 0.0 & 0.0 & 1.5 & 1.1 & 1.5 & 0.0 & 0.0 & 1.5 & 1.1 & 1.5\tabularnewline

	\midrule

\multicolumn{27}{c}{\textbf{ImageNet ResNet-152}}\tabularnewline

	\rowcolor{black!10}\cellcolor{white}\multirow{2}{*}{NSS~\citep{nss}} & DR & 2.9 & 19.1 & 39.6 & 47.2 & 41.6 & 2.9 & 19.9 & 39.6 & 46.5 & 41.1 & 4.2 & 31.2 & 41.4 & 9.1 & 32.9 & 1.1 & 12.6 & 28.3 & 35.7 & 29.1 & 1.0 & 11.9 & 29.8 & 33.3 & 28.7\tabularnewline
 & FPR & 0.4 & 1.4 & 1.2 & 1.4 & 2.0 & 0.4 & 1.4 & 1.2 & 1.4 & 2.0 & 0.4 & 1.4 & 1.2 & 1.4 & 2.0 & 0.6 & 1.4 & 1.2 & 1.4 & 2.0 & 0.4 & 1.4 & 1.2 & 1.4 & 2.0\tabularnewline

\rowcolor{black!10}\cellcolor{white}\multirow{2}{*}{ARC} & DR & 0.0 & 4.7 & 20.5 & \textbf{91.6} & 30.6 & 0.0 & 4.7 & 18.8 & \textbf{85.9} & 28.9 & 0.0 & 2.3 & 4.7 & \textbf{81.2} & 23.8 & 0.0 & 2.0 & 11.3 & \textbf{61.7} & 19.7 & 0.0 & 2.5 & 10.7 & \textbf{61.5} & 19.9\tabularnewline
 & FPR & 0.0 & 1.4 & 1.4 & 1.4 & 1.6 & 0.0 & 1.4 & 1.4 & 1.4 & 1.6 & 0.0 & 1.4 & 1.4 & 1.4 & 1.6 & 0.0 & 1.4 & 1.4 & 1.4 & 1.6 & 0.0 & 1.4 & 1.4 & 1.4 & 1.6\tabularnewline

\midrule

\multicolumn{27}{c}{\textbf{ImageNet SwinT-B-IN1K}}\tabularnewline

	\rowcolor{black!10}\cellcolor{white}\multirow{2}{*}{NSS~\citep{nss}} & DR & 4.5 & 16.2 & 42.4 & 47.5 & 44.2 & 4.9 & 15.8 & 41.8 & 47.1 & 44.1 & 12.3 & 28.7 & 29.3 & 4.5 & 28.9 & 1.6 & 11.0 & 31.3 & 35.5 & 31.1 & 1.4 & 10.4 & 31.8 & 35.1 & 30.8\tabularnewline
 & FPR & 0.6 & 1.0 & 1.2 & 1.6 & 2.3 & 0.6 & 1.0 & 1.2 & 1.6 & 2.3 & 0.6 & 1.0 & 1.2 & 1.5 & 2.3 & 0.6 & 1.0 & 1.2 & 1.6 & 2.3 & 0.6 & 1.0 & 1.2 & 1.6 & 2.3\tabularnewline

\rowcolor{black!10}\cellcolor{white}\multirow{2}{*}{ARC} & DR & 4.1 & 13.7 & \textbf{77.3} & \textbf{97.9} & 49.1 & 3.9 & 16.4 & \textbf{72.7} & \textbf{98.4} & 48.6 & 1.6 & 10.2 & \textbf{63.3} & \textbf{93.8} & 43.8 & 1.4 & 5.3 & 32.6 & \textbf{65.0} & 29.4 & 1.8 & 5.7 & 31.6 & \textbf{68.4} & 29.5\tabularnewline
 & FPR & 1.6 & 2.0 & 2.0 & 0.2 & 2.0 & 1.6 & 2.0 & 2.0 & 0.2 & 2.0 & 1.6 & 2.0 & 2.0 & 0.2 & 2.0 & 1.6 & 2.0 & 2.0 & 0.2 & 2.0 & 1.6 & 2.0 & 2.0 & 0.2 & 2.0\tabularnewline

 \bottomrule

\end{tabular}}

	\label{tab:sota}

\end{table}

As discussed in Sec.~\ref{sec:6}, due to our extremely limited problem setting
-- (1) no auxiliary deep model; (2) non-intrusive; (3) data-undemanding,
the most relevant methods that do not lack ImageNet evaluation are
\cite{nss,odds,convstat,safetynet,nic}.
But \cite{odds,convstat,safetynet,nic} still require a considerable amount
of data to build accurate (relatively) high-dimensional statistics.
The remaining NSS~\citep{nss} method craft
Natural Scene Statistics features, which are fed into SVM for binary classification.
We also adopt the trained SVMs in our ordinal regression framework,
with a reduced training set size to $100$ ($50$ benign + $50$ BIM adversarial)
for each SVM for a fair comparison.
All SVMs are tuned to control FPR.
The results and ROC curves for ``$\varepsilon{=}?$'' task can be found Tab.~\ref{tab:sota} and
Fig.~\ref{fig:roc}.
It is noted that
(1) SVM with the $18$-D NSS feature may fail to generalize due to insufficient sampling
(hence the below-diagonal ROC);
(2) NSS performs better for small $\varepsilon$, but performance
saturates with larger $\varepsilon$, because NSS does not incorporate any
cue from network gradient behavior;
(3) small $\varepsilon$ is difficult for ARC, but its performance soars with
larger $\varepsilon$ towards $100\%$, which is consistent and expected from our 
visualization;
(4) SVM with ARCv can generalize against all PGD-like attacks, while NSS failed for MIM;
(5) SVM with NSS may generalize against some non-PGD-like attacks~\citep{nss},
but not ARC due to SAE uniqueness;
(6) SVM with the $2$-D NSS feature (``Method 2'' in ~\cite{nss}) fails to generalize.
Thus, ARC achieves competitive performance consistently across different settings
despite the extreme limits, because
it is low-dimensional, and incorporates effective cues from gradients.
%

\subsection{Application of ARC: Attack Type Recognition}

By gathering the $14$ sets ($5$ sets from Tab.~\ref{tab:pgdlike} and $9$ sets from
Tab.~\ref{tab:nonpgdlike} t4-t12) of adversarial examples involved in
Tab.~\ref{tab:pgdlike} and Tab.~\ref{tab:nonpgdlike}, we can construct
a test dataset for attack type recognition.
Since each set has an equal number of samples, the binary classification
accuracy can be calculated as the average of the DR for PGD-like attacks
and $(1-\text{DR})$ for non-PGD-like attacks in the ``$\varepsilon=?$'' setting.
The results are $74.2\%$, $70.2\%$, $74.7\%$
for ResNet-18, ResNet-152, SwinT-B-IN1K, respectively.

\section{Discussions and Justifications}
\label{sec:5}

\begin{wrapfigure}{r}{0.32\textwidth}
	\centering
	\raisebox{0pt}[\dimexpr\height-0.6\baselineskip\relax]{%
		\includegraphics[width=0.30\textwidth]{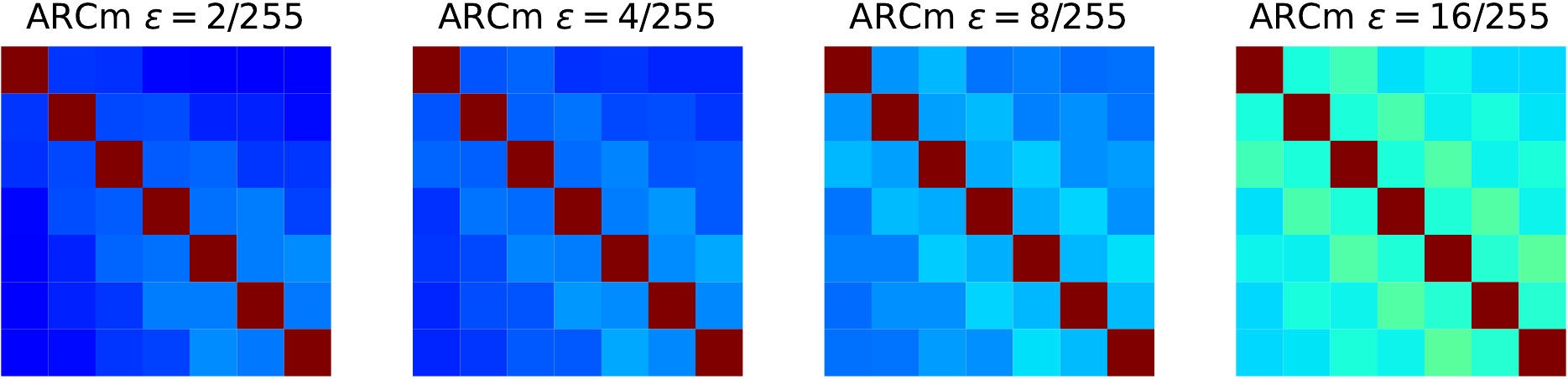}}\\
	\includegraphics[width=0.30\textwidth]{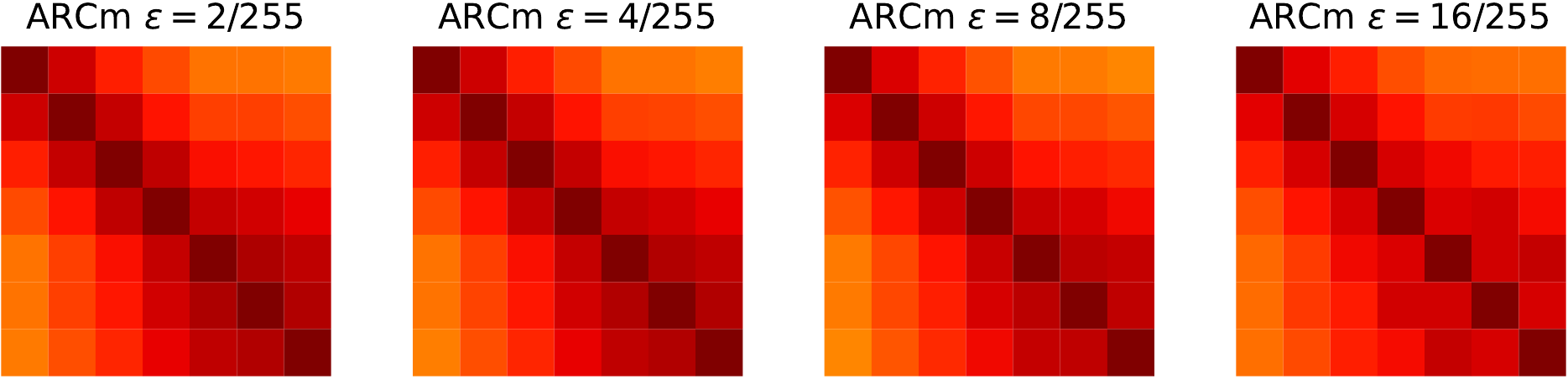}\\
	\includegraphics[width=0.30\textwidth]{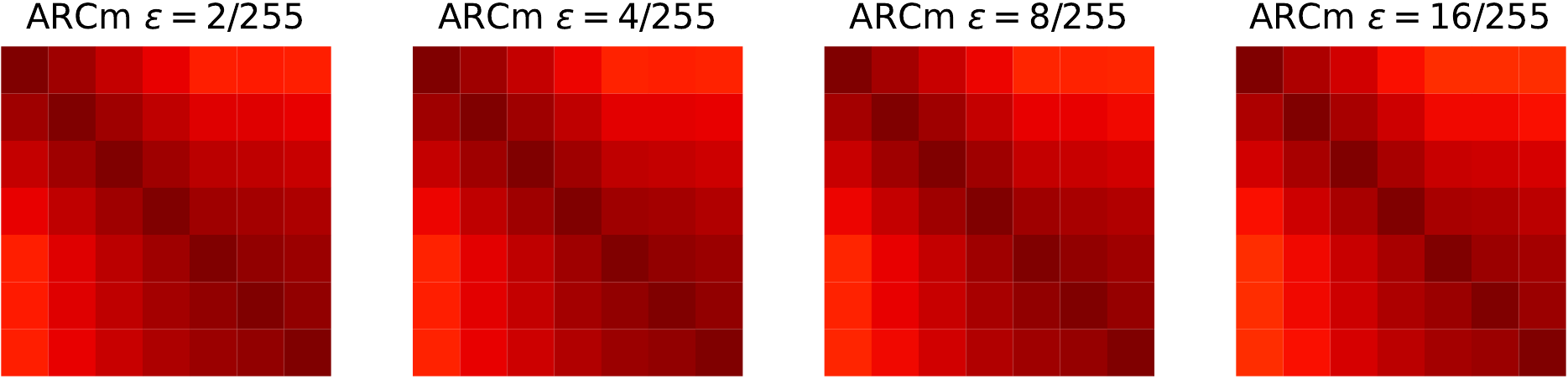}
	\caption{ARCm from regular (1\textsuperscript{st} row), and adversarially
	trained ResNet-50 (2\textsuperscript{nd} row w/ $\varepsilon{=}4/255$,
	3\textsuperscript{rd} row w/ $\varepsilon{=}8/255$).}
	\label{fig:defense}
\end{wrapfigure}
\textbf{Combination with Adversarial Training.}
From our experiment and recent works~\citep{madry,robustness,LLS}, its noted that
(1) small perturbations are hard to detect, but easy to defend; while
(2) large perturbations are hard to defend, but easy to detect.
However, combining defense and our detection is not effective on ImageNet.
As shown in Fig.~\ref{fig:defense}, we compute ARCm based on regular ResNet-50
(from PyTorch~\citep{pytorch})
and adversarially trained ResNet-50 on ImageNet (from \citep{robustness}).
Unlike the regular ResNet-50, adversarially trained one has a much higher
mean value in ARCm, resulting in almost non-separatable ARCv.
This means adversarial training makes the model very
linear around the data~\citep{operatornorm}.
Namely,
the network is trained to generalize while being already very linear
to the input, and thus it will be hard to make the model
behave even more linear to manipulate the output by the attack.

\textbf{Limitations.}
This paper focuses on characterizing a specific type of adversarial example
instead of a general detection or defense method. The following are the major limitations
of the ARC feature in potential applications:
(1) The SAE is unique but specific to PGD-like attacks;
(2) Jacobian computation is very slow for ImageNet models
because it requires $1000$ iterations of backward pass.
We are unable to evaluate our method on all ImageNet data with $2$ Nvidia Titan Xp GPUs.

\textbf{Further Discussions and justifications} involving ordinal regression,
training set size, and iterations of attacks can be found in the appendix.
Experiment details can be found in the supplementary code.

\section{Related Works}
\label{sec:6}

\textbf{Adversarial Attack \& Defense.}
Neural networks are found vulnerable~\citep{lbfgs,fgsm}.
Based on this, attacks with different threat models
are designed, including white-box attacks, transferability attacks,
and black-box attacks~\citep{benchmarking}.
\citet{bugfeature} attribute the existence of adversarial examples to non-robust
features.
To counter the attacks, adversarial training~\citep{madry,LLS,awp} is the most promising
defense, but it leads to an expensive training process and suffers from a notable generalization
gap.
Other types of defenses may suffer from adaptive attacks~\citep{obfuscated,adaptive}.

\textbf{Local Linearity} is revealed by~\citet{fgsm}, which leads to
a series of defenses and analysis.
\citet{LLS} regularize the model to behave linearly in the vicinity of data.
\citet{understandfast} show that the network being
highly non-linear locally results in FGSM training failure.
\citet{combating} suggest that a ``locally linear''
model can be used as a theoretical foundation for attacks and defenses.
%
         %
\citet{randomrelu} show that local linearity arises naturally at initialization.
Our method characterizes adversarial examples using local linearity.

\textbf{Adversarial Example Detection}~\citep{survey,carlini}
predicts whether a given image is adversarial or not.
This can be achieved through adversarial training~\citep{gat},
sub-network~\citep{subnetdetect} or extra loss~\citep{rce}, but
it will be costly for ImageNet.
Generative methods check reconstruction
error~\cite{magnet} or probability density~\cite{pixeldefend}, but are
data-demanding for accurate distributions.
Auxiliary deep models~\cite{dad,hgd} require a large amount of
data, and may suffer from adaptive attack~\citep{carlini}.
%
%
Feature statistics methods~\citep{convstat,odds,nss,safetynet,nic} 
leverage (high-dimensional) features, but most of them are still
data-demanding for an accurate statistics.
Many related works lack ImageNet evaluation and sensitivity analysis
with varying attack parameters even if the difficulty changes accordingly.

\section{Conclusions}
\label{sec:7}

We design an ARC feature with the
intuition that a model behaves more ``linear'' against adversarial examples
than benign ones, in which PGD-like attacks will leave a unique trace named SAE.
%
%
%


\bibliography{iclr2023_conference}
\bibliographystyle{iclr2023_conference}

\newpage
\appendix
\section{Additional Discussions}

\subsection{Summary of Pros \& Cons of the Proposed Method}

\textbf{Summary:}

	We design an ARC feature with the intuition that a model behaves more
	``linear'' against adversarial examples than benign ones, in which PGD-like
	attacks will leave a unique trace named SAE.
	ARC/SAE is adopted for two tasks
	to quantitatively demonstrate its effectiveness besides visualizations.

\textbf{Pros:}
\begin{itemize}[leftmargin=*]
	\item \underline{Intuitive.}
		The SAR and ARC are straightforward to interpret for human, since the
		meaning of the manually-crafted ARC features is clearly defined, and
		the feature dimensionality is low.
		Since SVM is employed for the two-dimensional ARCv, the whole pipeline
		of our method is intuitive and interpretable.
		This helps us to reveal more characteristics of adversarial examples.
	\item \underline{Light-weighted} in terms of algorithm components.
		Our method does not rely on any auxiliary deep model besides
		the model being attacked.
		It only involves hand-crafted ARC features and SVM.
		This makes our method usable for domains where it is hard to obtain
		or train an auxiliary deep model.
	\item \underline{Non-intrusive.}
		Our method does not require any change in the given neural network
		architecture or parameters.
		Instead, it analyzes the Jacobian matrices calculated from the neural network.
		This means the original model performance will not be affected.
	\item \underline{Data-undemanding.}
		Our method does not require a large number of training data like many
		other related works.
		Specifically, the simple cluster structure of the two-dimensional ARCv
		and SVM achieved such a low demand for data.
		This means our method is still usable in scenarios where it is impossible
		to access the full training dataset, such as Federated Learning.
	\item \underline{Unique to PGD-like attacks}.
		Our method is built upon strong assumptions, which makes it specific
		and only responds to PGD-like attacks. This characteristic
		can be used to identify the attack type and details in forensics scenarios.
		General attack detectors cannot do this because they cannot differentiate
		different types of attacks.
		This cannot be done by a general attack detector because they cannot
		differentiate different types of attacks.
		However, this point is meanwhile a disadvantage, see "Cons".
	\item \underline{Can infer attack details} such as loss function and the
		ground-truth labels, while most general attack detection methods cannot do the same.
	\item \underline{The stronger the attack is, the stronger the trace will be.}
		This is supported by the visualization results as well as the higher
		detection rate when $\varepsilon$ is larger.
		Previous methods compatible with our extremely-limited setting do not
		have such a property and may even perform worse with large perturbations
		in some cases (See Table 3).
		Other related works lack the attack parameter sensitivity analysis,
		whilst they can greatly change the difficulty to detect.
	\item \underline{Reveals a new perspective on understanding why Adversarial
		Training works.} See "Combination with Adversarial Training" in Section 5.
\end{itemize}

\textbf{Cons:}
\begin{itemize}[leftmargin=*]
	\item \underline{Is specific to PGD-like attacks} due to the strong assumptions.
		See "Uniqueness of SAE to PGD-Like Attack" in Section 2.
		It is not designed for non-PGD attacks since assumptions are broken.
		Ablation studies are carefully carried out in Section 4.2 to examine and
		justify these assumptions.
		Being specific to PGD-like attacks is meanwhile an advantage that leads
		to uniqueness (see "Pros").
	\item \underline{High time complexity due to Jacobian computation.}
		In practice, this is reflected by the time consumption of 
		the ARC feature calculation (See "Limitations" in Section 5).
		Experiments on ImageNet are extremely slow since calculating
		a single Jacobian matrix involves $1000$ (number of classes)
		times of backward pass of the neural network.
	\item Performs worse than the previous NSS method against \underline{small perturbations} (\emph{i.e.}, $\varepsilon=2/255$ or $\varepsilon=4/255$) for attack detection. (But significantly better against large perturbations).
	\item \underline{Incompatible with Adversarial Training.} But meanwhile it
		provides a new interpretation of why adversarial training works.
		See "Combination with Adversarial Training" in Section 5.
\end{itemize}

\subsection{Additional Discussiosn and Justifications}

\textbf{Ordinal Regression.}
Intuitively, the uninformed attack detection can be formulated as standard
regression to estimate a continuous $k$ value.
However, this introduces an undesired additional threshold hyper-parameter
for deciding whether an input with \emph{e.g.}, $0.5$ estimation is adversarial.
Ordinal regression produces discrete $k$ values and avoids such ambiguity and
unnecessary parameter.

\textbf{Training Set Size.}
Each of our SVMs has only $100$ training data (\emph{i.e.}, $50$ benign + $50$ adversarial).
The simple $2$-D ARCv distribution (Fig.~\ref{fig:arcfeat}) can be reflected by
a few data points, which even allows an SVM to generalize with less than $100$
data points (but may suffer from insufficient sampling with too few,
\emph{e.g.}, $10$+$10$ samples).
In contrast, the performance gain will be marginal starting from roughly
$200$ training samples, because the ARCv feature distribution is already well represented.

\textbf{Iterations of PGD-like Attacks}.
It is known that the number of iterations (fixed at $100$ in our experiments) also impacts the
attack strength besides perturbation magnitude $\varepsilon$.
\begin{wraptable}{r}{0.3\columnwidth}
\caption{Different iterations.}\label{tab:iterations}
	\resizebox{1.0\linewidth}{!}{%
\begin{tabular}{c|cccc}
	\toprule
Steps & DR & FPR & Acc & Acc{*}\tabularnewline
\midrule
100  & 79.2 & 1.1 & 0.0 & 62.4\tabularnewline
50  & 75.0  & 1.1  & 0.0  & 58.1\tabularnewline
25  & 64.1  & 1.1  & 0.0  & 47.3\tabularnewline
15  & 49.3  & 1.1 & 0.0  & 33.5\tabularnewline
10  & 33.1  & 1.1  & 0.2  & 20.1\tabularnewline
8  & 22.4  & 1.1  & 0.7  & 12.2\tabularnewline
5  & 7.1  & 1.1  & 3.7  & 5.5\tabularnewline
\bottomrule
\end{tabular}}
\end{wraptable}
As increasing number of iterations will also lead to a more linear response
from the model given a fixed and appropriate $\varepsilon$ and achieve
SAE similarly, we stick to one controlled variable $\varepsilon$ for simplicity.
On the contrary, reducing the number of iterations of a PGD-like attack
will also lead to small perturbations that are hard to detect (as demonstrated
in Section 4), and hence increase the possibility that the attack will not trigger
clear SAE and hence bypass the proposed detection method.
As an extreme case, FGSM, namely the single-step version of PGD does not effectively
trigger SAE (as discussed in Section 4.2).
The related works usually fix at a single set of attack parameters, and hence
miss the observation that smaller perturbations are harder to detect.
We conduct experiments with different numbers of steps of BIM attack on
CIFAR-10/ResNet-18, and report the corresponding results in Tab.~\ref{tab:iterations}.

\textbf{Future Recommendations.}
(1) Include ImageNet evaluation, as CIFAR-10 property may not hold on ImageNet;
(2) Check detector sensitivity \emph{w.r.t.} attack algorithm parameter,
as it may be significant.

\textbf{$T=48$ in Fig. 1.}
In our experiments, we use $T=6$, and the corresponding features are visualized
in Figure 2, Figure 3, and Figure 4. Some readers may want to know what the
feature will be like with a larger step size. Thus, we show this through T=48
examples in Figure 1, demonstrating that the network behaves more and more
linear from step to step. With an empirically chosen T=48, the trend of the
matrix is clear, and each cell in the matrix will not be too small to
visualize. With a larger step size like T=100 or even larger, the matrix will
show the same pattern, but the cells will be too small to visualize.

\subsection{Motivation of Extremely Limited Problem Setting}

An extremely limited problem setting (Paragraph 1 in Section 1) makes the proposed
method flexible and applicable in a wider range of scenarios
compared to existing methods.
Namely, a method can be used in more flexible scenarios
if it requires less from the adopter.

\textbf{Limited number of data samples.}
Data-demanding methods are only applicable for models with an accessible
training dataset.
In contrast, our method does not assume collecting a large amount of data
is easy for potential adopters.
Due to the low demand for data, the proposed method enables a wider
range of defense or forensics scenarios, including but not limited to
"Third-party Forensics" and "Federated Learning"
as follows:

\begin{itemize}[leftmargin=*]
	\item \underline{Third-party Forensics} (to identify whether a model is attacked,
		as well as infer attack details).
        Being data-undemanding means the proposed method can be applied to any
        pre-trained neural network randomly downloaded from the internet,
        or purchased from a commercial entity.
        For pre-trained neural networks using proprietary training datasets with
        commercial secret or ethic/privacy concerns
        (such as commercial face datasets and CT scans from patients),
		the proposed method is still
        valid as long as there are a few training samples for reference, or it is
        possible to request a few reference training samples.
	\item \underline{Federated Learning.}
        In federated learning, raw training data (such as face images)
        is forbidden to be transmitted to the central server due to user privacy.
		Even the neural network trainer cannot access the full training dataset
		(will violate user privacy), which makes data-demanding methods infeasible.
        In contrast, our proposed method is still valid in this scenario as
		long as a few (\emph{e.g.}, $50$) reference samples can be collected
		from volunteers.
\end{itemize}

\textbf{No change to network architecture or weights.}
Many models deployed in production are unaware of adversarial attacks.
Re-training and replacing these models will induce cost, and will even
introduce the risk of reducing benign example performance.

\textbf{No auxiliary deep networks.}
Since a large amount of data is assumed to be not easy to obtain due to
commercial or ethic reasons, training auxiliary deep networks is not always feasible.
Pre-trained auxiliary deep networks are not always available for classification in any domain.

\subsection{More on Adaptive Attack}

According to \cite{adaptive}, some similar attack detection methods
are broken by adaptive attacks. Here we discuss more about the existing
adaptive attacks and report the quantitative experimental results.
We also further elaborate on the adaptive attack mentioned in Section 2.

\textbf{Logit Matching.} (from Section 5.2 "The Odds are Odd" of \cite{adaptive})
Instead of maximizing the default entropy loss, we switch to minimize the
MSE loss between the clean logits from another class and that of the
adversarial example.
We conduct the experiment with CIFAR-10 and ImageNet following the setting in Section 4.
The experimental results can be found in the following table.
Note, switching the loss function to MSE (Logit Matching) breaks our condition (IV).
However, the attack still triggers SAE through the least-likely class,
and hence our method is still effective, but is (expectedly) weaker
than the BIM with the original cross-entropy loss.

\begin{table}[h]
    \resizebox{1.0\linewidth}{!}{%
\begin{tabular}{cc|cccc|cccc|cccc|cccc|cccc}

	\toprule

\multirow{2}{*}{\textbf{\thead{Dataset\\Model}}} & \multirow{2}{*}{\textbf{Attack}} & \multicolumn{4}{c|}{$\epsilon=2/255$} & \multicolumn{4}{c|}{$\epsilon=4/255$} & \multicolumn{4}{c|}{$\epsilon=8/255$} & \multicolumn{4}{c|}{$\epsilon=16/255$} & \multicolumn{4}{c}{$\epsilon=?$}\tabularnewline
\cline{3-22} \cline{4-22} \cline{5-22} \cline{6-22} \cline{7-22} \cline{8-22} \cline{9-22} \cline{10-22} \cline{11-22} \cline{12-22} \cline{13-22} \cline{14-22} \cline{15-22} \cline{16-22} \cline{17-22} \cline{18-22} \cline{19-22} \cline{20-22} \cline{21-22} \cline{22-22} 
 &  & DR & FPR & Acc & Acc{*} & DR & FPR & Acc & Acc{*} & DR & FPR & Acc & Acc{*} & DR & FPR & Acc & Acc{*} & DR & FPR & Acc & Acc{*}\tabularnewline

	\midrule

\multirow{1}{*}{\makecell{CIFAR-10\\ResNet-18}} & BIM (Logit Matching) & 0.0 & 0.0 & 80.6 & 80.6 & 0.0 & 0.0 & 63.2 & 63.2 & 23.8 & 1.5 & 46.3 & 35.5 & 48.0 & 1.1 & 38.0 & 20.2 & 22.8 & 1.5 & 57.1 & 46.9\tabularnewline
\tabularnewline
\hline
    \multirow{1}{*}{\makecell{ImageNet\\ResNet-152}} & BIM (Logit Matching) & 0.0 & 0.0 & 46.1 & 46.1 & 7.0 & 1.4 & 18.8 & 17.2 & 17.2 & 1.4 & 9.4 & 7.0 & \textbf{91.4} & 1.4 & 3.1 & 0.0 & 30.3 & 1.6 & 19.3 & 17.6\tabularnewline
\tabularnewline
\hline
    \multirow{1}{*}{\makecell{ImageNet\\SwinT-B-IN1K}} & BIM (Logit Matching) & 0.8 & 1.6 & 46.1 & 45.3 & 7.0 & 2.0 & 7.0 & 7.0 & \textbf{55.5} & 2.0 & 0.8 & 0.8 & \textbf{90.6} & 0.2 & 0.0 & 0.0 & 41.2 & 2.0 & 13.5 & 13.1\tabularnewline
\tabularnewline

\bottomrule

\end{tabular}}
    \caption{Results of Logit Matching as adaptive attack against our method.}
\end{table}

\textbf{Interpolation with Binary Search.}
(from Section 5.13 "Turning a Weakness into a Strength" of \cite{adaptive})
This method finds interpolated adversarial examples close to the
decision boundary with binary search.
We conduct the experiment with CIFAR-10 and ImageNet.
The experimental results can be found in the following table.
Compared to the baseline results, the results show that our method is still
effective against the adversarial examples close to the decision boundary.

\begin{table}[h]
    \resizebox{1.0\linewidth}{!}{%
\begin{tabular}{cc|cccc|cccc|cccc|cccc|cccc}

	\toprule

\multirow{2}{*}{\textbf{\thead{Dataset\\Model}}} & \multirow{2}{*}{\textbf{Attack}} & \multicolumn{4}{c|}{$\epsilon=2/255$} & \multicolumn{4}{c|}{$\epsilon=4/255$} & \multicolumn{4}{c|}{$\epsilon=8/255$} & \multicolumn{4}{c|}{$\epsilon=16/255$} & \multicolumn{4}{c}{$\epsilon=?$}\tabularnewline
\cline{3-22} \cline{4-22} \cline{5-22} \cline{6-22} \cline{7-22} \cline{8-22} \cline{9-22} \cline{10-22} \cline{11-22} \cline{12-22} \cline{13-22} \cline{14-22} \cline{15-22} \cline{16-22} \cline{17-22} \cline{18-22} \cline{19-22} \cline{20-22} \cline{21-22} \cline{22-22} 
 &  & DR & FPR & Acc & Acc{*} & DR & FPR & Acc & Acc{*} & DR & FPR & Acc & Acc{*} & DR & FPR & Acc & Acc{*} & DR & FPR & Acc & Acc{*}\tabularnewline

	\midrule

\multirow{1}{*}{\makecell{CIFAR-10\\ResNet-18}} & BIM (Interpolation) & 0.0 & 0.0 & 65.7 & 65.7 & 0.0 & 0.0 & 44.6 & 44.6 & 28.0 & 1.5 & 21.9 & 28.0 & \textbf{74.4} & 1.1 & 6.0 & 56.4 & 28.0 & 1.5 & 34.6 & 48.8\tabularnewline
\tabularnewline
\hline
    \multirow{1}{*}{\makecell{ImageNet\\ResNet-152}} & BIM (Interpolation) & 0.0 & 0.0 & 18.8 & 18.8 & 4.7 & 1.4 & 6.2 & 5.5 & 25.0 & 1.4 & 0.8 & 0.8 & \textbf{90.6} & 1.4 & 0.0 & 0.8 & 31.4 & 1.6 & 6.4 & 6.2\tabularnewline
\tabularnewline
\hline
    \multirow{1}{*}{\makecell{ImageNet\\SwinT-B-IN1K}} & BIM (Interpolation) & 1.6 & 1.6 & 44.5 & 45.3 & 3.9 & 2.0 & 37.5 & 35.9 & \textbf{66.4} & 2.0 & 14.1 & 64.8 & \textbf{97.7} & 0.2 & 0.0 & 97.7 & 42.8 & 2.0 & 24.0 & 61.3\tabularnewline
\tabularnewline

\bottomrule

\end{tabular}}
    \caption{Results of Interpolation with Binary Search as adaptive attack against our method.}
\end{table}

\textbf{Adaptive Attack discussed in Section 2.} To avoid triggering SAE, the
goal of the PGD attack can include an additional term to minimize $\|\bm{S}_*(\bm{x}+\bm{r})\|_F$.
Namely, the corresponding adaptive attack is:
\begin{align*}
    & \arg\max_{\bm{r}}L_{\text{CE}}(\bm{x}+\bm{r},c(\bm{x}))-\|\bm{S}_{*}(\bm{x}+\bm{r})\|_F\\
    = & \arg\max_{\bm{r}}L_{\text{CE}}(\bm{x}+\bm{r},c(\bm{x}))-[\sum_{i}\sum_{j}|s_{n^{*}}^{(i,j)}|^{2}]^{1/2}\\
    = & \arg\max_{\bm{r}}L_{\text{CE}}(\bm{x}+\bm{r},c(\bm{x}))-[\sum_{i=1}^{T+1}\sum_{j=1}^{T+1}\cos[\nabla f_{n^{*}}(\bm{x}+\bm{r}+\bm{\delta}_{i}),\nabla f_{n^{*}}(\bm{x}+\bm{r}+\bm{\delta}_{j})]^{2}]^{1/2}
\end{align*}

To solve this adaptive attack problem, the straightforward solution is to conduct
$Z$-step PGD updates with the modified loss function.
Each step includes but is not limited to these computations: (1) $T+1$
Jacobian matrices to calculate $n^{*}$ and $\nabla f_{n^{*}}(\cdot)$;
(2) $T+1$ Hessian matrices to calculate $\nabla^{2}f_{n^{*}}(\cdot)$.
Let $\psi_{J}$ and $\psi_{H}$ be the time consumption for Jacobian
and Hessian matrices respectively. Then the time consumption of the
$Z$ steps of optimization in total is greater than $Z(T+1)(\psi_{J}+\psi_{H})$.

For reference, for Nvidia Titan Xp GPU and CIFAR-10/ResNet-18, the
$\psi_{J}=0.187\pm0.012$ seconds, and $\psi_{H}=20.959\pm0.679$
seconds (Python code for this benchmark can be found in Appendix).
If we use $Z=100$ steps of PGD attack, and $T=6$ for calculating
ARC, each adversarial example of a CIFAR-10 image takes more than $Z(T+1)(\psi_{J}+\psi_{H})\approx14802$
seconds (i.e., $4.1$ hours).

Note, we acknowledge that other alternative adaptive attack designs are
possible. However, as long as the alternative design involves optimizing any loss
term calculated from gradients, second-order gradients (Hessian) will be
required to finish the optimization process, which again makes the alternative
attack computationally prohibitive.
Switching to non-PGD-like attacks is much simpler.

\subsection{More on Related Works (Extension to Section 6)}

\textbf{Defenses with similar ideas.}
\begin{itemize}[leftmargin=*]
    \item ``The Odds are Odd''~\cite{odds} is an attack detection method based on
        a feature statistic test. This method is categorized in Section 6 as feature
        statistics-based methods.
        In particular, it detects adversarial examples based on the difference between
        the logits of clean images and images with random noise.
        This method assumes that a random noise may break the adversarial perturbation
        and hence lead to notable changes in the logits, and it
        is capable of correcting test time predictions.
        Meanwhile, it can be broken by the adaptive attack to match the logits with
        an image from another example~\cite{adaptive}.
        Similarly, our method can be seen as a statistical test for gradient
        consistency as reflected by the ARC feature.
        Our method is motivated by the assumption that neural networks will manifest
        ``local linearity'' with respect to adversarial examples, which will not happen
        for benign examples.
        Meanwhile the SAE is consistent
        across different architectures, and the corresponding 2-D ARCv feature shows
        a very simple cluster structure for both benign and adversarial examples.
        The adaptive attack against \cite{odds} can merely slightly reduce
        the effectiveness of our attack, as shown in the additional adaptive
        attack experiments in this Appendix.
\item ``Turning a Weakness into a Strength''~\cite{turnweakness} is an attack
    detection method that is conceptually similar to \cite{odds}.
        This method involves two criteria for detection: (1) low density of
        adversarial perturbations -- random perturbations applied to natural
        images should not lead to changes in the predicted label.
        The input will be rejected if the change in the predicted probability
        vector is significant after adding Gaussian noise.
        (2) close proximity to the decision boundary -- this leads to a method
        that rejects an input if it requires too many steps to successfully
        perturb with an iterative attack algorithm.
        Hence, this method can be seen as a detector with a two-dimensional
        manually crafted feature.
        This method can be broken by an adaptive attack~\cite{adaptive}
        that searches for interpolation between the benign and adversarial
        examples.
        Similarly, our method leverages BIM, an iterative attack to calculate
        the ARC feature.
        However, differently, our method uses the iterative attack to explore
        the local area around the input, in order to calculate the extent
        of ``local linearity'' around the point as the ARC feature, while
        \cite{turnweakness} leverage an iterative attack to count the 
        number of required steps.
        The ARC feature shows a clear difference between benign and adversarial
        examples, and hence does not need to combine with other manually
        crafted features.
        \cite{turnweakness} point out that solely using one criterion is
        insufficient, because criterion (1) may be easily bypassed.
        The adaptive attack against \cite{turnweakness} can merely slightly reduce
        the effectiveness of our attack, as shown in the additional adaptive
        attack experiments in this Appendix.
\end{itemize}

\subsection{Python Code for Evaluating Time Consumption of Jacobian / Hessian}

The python code for measuring the time consumption for Jacobian and Hessian matrices
calculation is shown below. The code is based on CIFAR-10 settings with $M=3\times 32\times 32$
and $N=10$, and the neural network used is ResNet-18.
For reference, the result on Nvidia Titan Xp GPU
is $0.187\pm 0.012$ seconds for Jacobian, and $20.959\pm 0.679$ seconds for Hessian.

Note, for the ImageNet/ResNet-152 case, the Jacobian and Hessian calculation
cost is much higher.

\begin{framed}
\scriptsize
\input{benchmark.tex}
\end{framed}

\end{document}